\documentclass{article}

\usepackage{iclr2027_conference,times}

\usepackage{amsmath,amsfonts,bm}

\def\eqref#1{equation~\ref{#1}}

\def\plaineqref#1{\ref{#1}}

\def\1{\bm{1}}

\DeclareMathAlphabet{\mathsfit}{\encodingdefault}{\sfdefault}{m}{sl}
\SetMathAlphabet{\mathsfit}{bold}{\encodingdefault}{\sfdefault}{bx}{n}

\usepackage{hyperref}
\usepackage{url}

\usepackage[expansion=false]{microtype}
\usepackage{graphicx}
\graphicspath{{images/}}
\usepackage{subcaption}
\usepackage{booktabs}
\usepackage{xcolor}
\usepackage{algorithm}
\usepackage{algorithmic}
\usepackage{amsmath}
\usepackage{amssymb}
\usepackage{mathtools}

\title{Q-Drift: Quantization-Aware Drift Correction for Diffusion Model Sampling}

\author{Sooyoung Ryu$^{1}$ \quad Mathieu Salzmann$^{2}$ \quad Saqib Javed$^{2}$ \\
$^{1}$Seoul National University \quad $^{2}$EPFL \\
\texttt{ryuswimming@snu.ac.kr}}

\iclrfinalcopy

\begin{document}

\maketitle
\lhead{Preprint}

\begin{abstract}
Post-training quantization (PTQ) is a practical path to deploy large diffusion models, but quantization noise can accumulate over the denoising trajectory and degrade generation quality.
We propose Q-Drift, a sampler-side correction that aims to preserve the intended sampling marginals through a deterministic drift adjustment motivated by generalized marginal-preserving SDEs.
Q-Drift uses the calibrated conditional residual variance of quantization error to determine a correction factor at each step.
Our SDXL study shows that calibration with as few as 10 paired full-precision/quantized runs remains effective.
The resulting sampler correction is plug-and-play with common samplers, diffusion models, and PTQ methods, while incurring negligible overhead at inference.
Across six diverse text-to-image models (spanning DiT and U-Net), three samplers (Euler, flow-matching, DPM-Solver++), and two PTQ methods (SVDQuant, MixDQ), Q-Drift improves FID over the corresponding quantized baseline in all seven settings of our main evaluation, with up to 4.79 FID reduction on PixArt-$\Sigma$ (SVDQuant W3A4), while preserving CLIP scores.
Code is available at \url{https://github.com/sooyoung-ryu/Q-Drift}.
\end{abstract}

\section{Introduction}

Diffusion models have become a dominant paradigm for high-quality image generation
\citep{ho2020ddpm, song2021score}, and modern text-to-image systems such as Stable Diffusion have shown their
practical effectiveness in real-world generation \citep{rombach2022high, podell2024sdxl}.
Their iterative denoising and growing scale, however, incur substantial computational and memory costs, which pose
significant challenges for deployment in resource-constrained or latency-critical settings
\citep{shen2025efficient, karras2022elucidating}.

Post-training quantization (PTQ) converts a pretrained model into a low-bit representation using only a small
calibration set, avoiding the training cost of quantization-aware training (QAT), and it has become a practical and
widely adopted way to compress diffusion models.
Nevertheless, quantization-induced degradation remains a persistent challenge, especially under aggressive low-bit
settings or with fast, few-step samplers
\citep{shang2023ptq4dm, li2023qdiffusion, he2023ptqd}.
Recent work mitigates it by improving the fidelity of the quantized model to its full-precision counterpart:
SVDQuant absorbs extreme outliers into low-rank components \citep{li2025svdquant}, and MixDQ uses
metric-decoupled sensitivity analysis with integer-programming-based mixed-precision allocation \citep{zhao2024mixdq}.
These approaches substantially reduce model-level quantization error, yet noticeable degradation in generation
quality often persists in practice
\citep{he2023ptqd, yao2024tac}.

Many PTQ methods optimize layer-wise reconstruction at individual sampling steps.
These objectives do not directly control the accumulated sampling error.
Generation, however, is iterative: the denoiser output at one step determines the input to the next, so small local
errors accumulate along the trajectory \citep{li2024errorprop}, and quantization error therefore interacts with the
sampling process rather than remaining purely step-local.

Motivated by this, we introduce \emph{Quantization-Aware Drift Correction (Q-Drift)}, a sampler-side
method that approaches quantized diffusion sampling from a distributional perspective and aims to preserve
the intended sampling marginals. The key idea is to use the conditional residual variance of quantization error
to determine a deterministic drift adjustment, drawing on the drift--diffusion pairing in generalized
marginal-preserving SDEs \citep{song2021score,karras2022elucidating}.
This connects a statistic available from calibration to a correction of the sampling dynamics.

Q-Drift translates this design into a lightweight correction: one precomputed scalar factor per sampling
step. It complements model-level PTQ methods such as SVDQuant and MixDQ~\citep{li2025svdquant,zhao2024mixdq},
requires no retraining or weight changes, and incurs negligible inference overhead.
The same rescaling principle supports adaptations to Euler, flow-matching, and DPM-Solver++ samplers.
Our SDXL study further shows that as few as 10 paired full-precision/quantized calibration runs suffice to retain
the correction's effectiveness.

We evaluate Q-Drift on six text-to-image models spanning DiT and U-Net architectures, three samplers,
and two PTQ methods. It improves FID in all seven settings of our main evaluation at comparable CLIP scores,
with a reduction of up to 4.79 FID on PixArt-$\Sigma$.
Our SDXL ablations support conditional residual variance and a single scalar factor per step as practical
design choices. Together, these results demonstrate that a lightweight sampler-side adjustment can improve
distributional fidelity across diverse quantized diffusion models.

\section{Related Work}

\subsection{Diffusion Sampling and SDE Perspectives}
Diffusion models generate images by iteratively denoising from a simple Gaussian and have become a central paradigm for high-fidelity generation~\citep{sohl2015deep, ho2020ddpm, nichol2021improved, dhariwal2021diffusion}.
A continuous-time formulation connects sampling to reverse-time SDEs and the probability-flow ODE~\citep{anderson1982reverse, haussmann1986time, song2019score, song2021score}, providing a principled lens on discretization, schedules, and the accuracy--efficiency tradeoff at inference~\citep{karras2022elucidating, shen2025efficient}, and it has driven extensive work on retraining-free acceleration through better discretization and higher-order ODE/SDE solvers~\citep{song2020ddim, liu2022pndm, lu2022dpm_solver, zhang2023deis, zhao2023unipc, lu2023dpm_solver_v3, xue2023sasolver} as well as distillation into few-step generators and consistency mappings~\citep{salimans2022progressive, song2023consistency, luo2023lcm, sauer2024add, yin2024idd}.
While these lines of work substantially reduce the number of function evaluations (NFEs), deployment still faces significant memory and compute costs for large diffusion backbones~\citep{rombach2022high, peebles2023dit}, motivating the complementary compression techniques discussed below.

\subsection{Quantizing Diffusion Models}
PTQ is a standard approach for compressing large generative models and is particularly attractive for diffusion models because it avoids retraining~\citep{adaround2020, shang2023ptq4dm, li2023qdiffusion}.
A key challenge is that activation and output statistics shift substantially across denoising timesteps, and sequential generation can amplify small approximation errors, making calibration and distribution alignment central to maintaining generation quality~\citep{shang2023ptq4dm, li2023qdiffusion, liu2024edadm, wang2024apqdm, li2023qdm, he2023ptqd, huang2024tfmq, chu2024qncd, li2024errorprop}.
Accordingly, many PTQ methods adapt reconstruction-based quantization to the multi-timestep setting and combine it with diffusion-specific calibration or mixed-precision choices~\citep{li2021brecq, shang2023ptq4dm, li2023qdiffusion, wang2024apqdm, sun2024tmpqdm}, while others target temporal and conditioning pathways or modality-specific backbones~\citep{huang2024tfmq, tian2024qvd}.
For text-to-image diffusion, additional sensitivity in cross-attention and conditioning modules motivates progressive calibration, activation relaxation, metric-driven mixed precision, and distribution-aware grouping~\citep{pandey2023softmaxbias, tang2024pcr, zhao2024mixdq, ryu2025dgq}, and when PTQ quality drops at aggressive bitwidths, QAT or selective fine-tuning becomes increasingly important~\citep{he2023efficientdm, wang2024quest, ryu2024meft, guo2024intlora}.
As backbones have moved from U-Nets to transformers (DiTs), quantization research has extended to transformer-specific sensitivity analyses and PTQ pipelines built for timestep-dependent behaviors and structured outliers~\citep{peebles2023dit, bao2023uvit, yang2024aqdit, wu2024ptq4dit, chen2025qdit, dong2024ditas, liu2024hqdit}, with aggressive settings (e.g., 4-bit) increasingly emphasizing outlier- and distribution-aware quantization~\citep{li2025svdquant, lee2025dmq, zhao2025viditq}.

\subsection{Quantization Noise Compensation During Sampling}
\label{sec:related:sampling}
Since diffusion sampling is a sequential procedure where each denoiser output becomes the next-step input, small approximation errors can propagate and amplify across timesteps \citep{li2024errorprop}.
For quantized diffusion models, a growing body of work explicitly models quantization noise during sampling and proposes step- or timestep-dependent correction mechanisms.
PTQD~\citep{he2023ptqd} decomposes quantization noise into correlated and residual components and compensates the induced mean/variance mismatch in the denoising process.
TAC-Diffusion~\citep{yao2024tac} proposes a timestep-aware correction that dynamically adjusts quantization errors during inference.
QNCD~\citep{chu2024qncd} estimates the accumulated quantization noise at run time and filters it out during sampling.
D$^{2}$-DPM~\citep{zeng2025d2dpm} introduces a dual denoising mechanism to mitigate quantization-induced noise accumulation during iterative sampling.
We compare these methods and our method in Sec.~\ref{sec:exp:sampler_side}.

Our work is most closely related to sampler-time compensation approaches, but differs in objective and mechanism.
Our objective is to preserve the intended sampling marginals. We use calibrated conditional residual variance
to determine a deterministic drift rescaling motivated by generalized marginal-preserving SDEs.

\section{Preliminaries}
\label{sec:prelim}

\subsection{Diffusion Sampling as a Generalized SDE}
\label{sec:prelim:sde}
Diffusion models generate samples by simulating a reverse-time stochastic process. This process is
designed so that its marginal distribution $p_t(\mathbf{x})$ at each noise level matches that of a fixed
forward noising process \citep{sohl2015deep,ho2020ddpm,song2021score}.
Following the noise-level parameterization of \citet{karras2022elucidating}, we write
$p_t(\mathbf{x})\equiv p(\mathbf{x};\sigma(t))$ with $\dot\sigma(t)>0$; sampling runs backward in $t$, so noise decreases.

Following \citet{karras2022elucidating}, we use the generalized reverse-time SDEs below, which share the same
target marginals under the exact score in continuous time:
\begin{equation}
d\mathbf{x}
=
\underbrace{-\dot{\sigma}(t)\sigma(t)\nabla_{\mathbf{x}}\log p(\mathbf{x};\sigma(t))\,dt}_{\text{probability-flow ODE}}
\underbrace{
\underbrace{-\beta(t)\sigma(t)^2\nabla_{\mathbf{x}}\log p(\mathbf{x};\sigma(t))\,dt}_{\text{deterministic noise decay}}
\underbrace{+\sqrt{2\beta(t)}\,\sigma(t)\,d\mathbf{w}_t}_{\text{noise injection}}
}_{\text{Langevin diffusion SDE}},
\label{eq:general_sde_beta}
\end{equation}
where $\mathbf{w}_t$ denotes a reverse-time Wiener process with increment covariance $|dt|\mathbf{I}$.
In the Langevin diffusion SDE term, $\beta(t)\geq 0$ freely controls the amount of stochasticity by pairing a
deterministic noise decay (Langevin drift) with fresh noise injection (Langevin diffusion),
while preserving the same marginals.
This pairing motivates the deterministic correction used by Q-Drift.

\subsection{Euler--Maruyama Discretization}
\label{sec:prelim:em}
For an It\^{o} SDE $d\mathbf{x}=\mathbf{a}(\mathbf{x},t)\,dt+\mathbf{b}(t)\,d\mathbf{w}_t$ on a time grid $\{t_i\}$
with $\Delta t_i\triangleq t_{i+1}-t_i$, the Euler--Maruyama discretization~\citep{kloeden1992numerical} gives
\begin{equation}
\mathbf{x}_{i+1}
=\, \mathbf{x}_i
+ \mathbf{a}(\mathbf{x}_i,t_i)\,\Delta t_i
+ \mathbf{b}(t_i)\sqrt{|\Delta t_i|}\,\mathbf{z}_i ,
\qquad \mathbf{z}_i \sim \mathcal{N}(\mathbf{0},\mathbf{I}),
\label{eq:em_generic}
\end{equation}
so the drift scales with $\Delta t_i$ and the diffusion with $\sqrt{|\Delta t_i|}$.

\subsection{Empirical Joint Gaussianity of Quantization Noise and Output}
\label{sec:prelim:joint-gauss}
We follow the empirical characterization of quantization effects reported in D$^2$-DPM~\citep{zeng2025d2dpm}.
At a fixed timestep (equivalently, a fixed noise level), let $\epsilon_{\theta}^{(t)}$ denote the full-precision
noise prediction and $\hat{\epsilon}_{\theta}^{(t)}$ its quantized counterpart, and define the
quantization noise as
$\Delta\epsilon_{\theta}^{(t)} \triangleq \hat{\epsilon}_{\theta}^{(t)} - \epsilon_{\theta}^{(t)}$. When the
timestep is clear we drop the superscript and write $\Delta\epsilon$, or $\Delta\epsilon_i$ at step $i$.
D$^2$-DPM reports that at each timestep both $\Delta\epsilon_{\theta}^{(t)}$ and
$\hat{\epsilon}_{\theta}^{(t)}$ are well approximated by Gaussians, that the two are clearly correlated, and
that their joint density is approximately elliptical.
Accordingly, D$^2$-DPM models the two jointly as
\begin{equation}
\begin{bmatrix}
\hat{\epsilon}_{\theta}^{(t)}\\
\Delta\epsilon_{\theta}^{(t)}
\end{bmatrix}
\sim
\mathcal{N}\!\left(
\begin{bmatrix}
\mu_{\hat{\epsilon}}^{(t)}\\
\mu_{\Delta}^{(t)}
\end{bmatrix},
\begin{bmatrix}
\Sigma_{\hat{\epsilon}\hat{\epsilon}}^{(t)} & \Sigma_{\hat{\epsilon}\Delta}^{(t)}\\
\Sigma_{\Delta\hat{\epsilon}}^{(t)} & \Sigma_{\Delta\Delta}^{(t)}
\end{bmatrix}
\right).
\label{eq:joint_gaussian_d2}
\end{equation}
Under the joint Gaussian model in Eq.~\plaineqref{eq:joint_gaussian_d2}, the conditional distribution
$\Delta\epsilon_{\theta}^{(t)} \mid \hat{\epsilon}_{\theta}^{(t)}=\hat{\epsilon}$ is Gaussian, i.e.,
\begin{equation}
\Delta\epsilon_{\theta}^{(t)} \mid \hat{\epsilon}_{\theta}^{(t)}=\hat{\epsilon}
\sim
\mathcal{N}\!\bigl(\mu_{\mathrm{cond}}^{(t)}(\hat{\epsilon}), \Sigma_{\mathrm{cond}}^{(t)}\bigr),
\label{eq:cond_gaussian_d2}
\end{equation}
where the conditional mean and covariance admit closed-form expressions:
\begin{equation}
\mu_{\mathrm{cond}}^{(t)}(\hat{\epsilon}) = \mu_{\Delta}^{(t)} + \Sigma_{\Delta\hat{\epsilon}}^{(t)}\bigl(\Sigma_{\hat{\epsilon}\hat{\epsilon}}^{(t)}\bigr)^{-1}\bigl(\hat{\epsilon}-\mu_{\hat{\epsilon}}^{(t)}\bigr),
\label{eq:cond_mean_d2}
\end{equation}
\begin{equation}
\Sigma_{\mathrm{cond}}^{(t)} = \Sigma_{\Delta\Delta}^{(t)} - \Sigma_{\Delta\hat{\epsilon}}^{(t)}\bigl(\Sigma_{\hat{\epsilon}\hat{\epsilon}}^{(t)}\bigr)^{-1}\Sigma_{\hat{\epsilon}\Delta}^{(t)}.
\label{eq:cond_cov_d2}
\end{equation}

\subsection{Practical Simplifications for Joint Gaussian Modeling}
\label{sec:prelim:joint-simplify}
Estimating the full covariance blocks in Eq.~\plaineqref{eq:joint_gaussian_d2} is impractical at this
dimensionality, so, following D$^2$-DPM~\citep{zeng2025d2dpm}, we use diagonal covariance blocks (retaining the
per-element correlation between $\hat{\epsilon}_{\theta}^{(t)}$ and $\Delta\epsilon_{\theta}^{(t)}$ through a
diagonal cross-covariance), an isotropic parameterization of each block, and separate statistics per timestep,
since quantization effects vary across noise levels.
These choices make the model estimable from a small calibration set.
Appendix~\ref{sec:app:assumption_validation} examines selected distributional and covariance properties
underlying these approximations. For the local variance-matching construction in
Sec.~\ref{sec:method:derivation}, we use independent auxiliary perturbations across steps.

\section{Method}
\label{sec:method}

\subsection{Overview}
\label{sec:method:overview}
Q-Drift aims to preserve the intended sampling marginals under quantization.
We use the empirical Gaussian modeling of quantization error and its correlation with the quantized output
(Sec.~\ref{sec:prelim:joint-gauss}--\ref{sec:prelim:joint-simplify}) to estimate a timestep-wise conditional
residual variance. This statistic determines the correction scale through the drift--diffusion pairing in
Eq.~\plaineqref{eq:general_sde_beta}. The resulting method rescales the deterministic update with one calibrated
factor per step, without retraining.

\subsection{Derivation for a First-Order Euler Sampler}
\label{sec:method:derivation}

We derive the correction scale by (i) identifying the quantization perturbation in the Euler update,
(ii) representing its calibrated residual variability with an auxiliary Gaussian perturbation, and
(iii) matching the auxiliary variance to the SDE diffusion term and using the paired drift.

\paragraph{$\sigma$-parameterized generalized marginal-preserving SDEs.}
Starting from Eq.~\plaineqref{eq:general_sde_beta} and using the noise-level parameterization $\sigma=\sigma(t)$ with
$d\sigma=\dot{\sigma}(t)\,dt$, we can reparameterize the reverse-time dynamics in $\sigma$ as
\begin{equation}
d\mathbf{x} = -\sigma\,\nabla_{\mathbf{x}}\log p(\mathbf{x};\sigma)\,d\sigma - \tilde{\beta}_{\sigma}(\sigma)\,\sigma^2\,\nabla_{\mathbf{x}}\log p(\mathbf{x};\sigma)\,d\sigma + \sigma\sqrt{2\tilde{\beta}_{\sigma}(\sigma)}\,d\mathbf{w}_{\sigma},
\label{eq:method:sde_sigma}
\end{equation}
where $\tilde{\beta}_{\sigma}(\sigma)\triangleq \beta(t)/\dot{\sigma}(t)\ge 0$, and $\mathbf{w}_{\sigma}$ is a
Wiener process indexed by $\sigma$ (so $\mathrm{Var}(d\mathbf{w}_{\sigma})=|d\sigma|$).
As in Eq.~\plaineqref{eq:general_sde_beta}, $\tilde{\beta}_{\sigma}$ is a free parameter that simultaneously controls
the injected noise (Langevin diffusion) and the paired deterministic noise decay (Langevin drift), preserving the
same marginals \citep{karras2022elucidating}.

\paragraph{Euler sampler from the probability-flow ODE.}
Setting $\tilde{\beta}_{\sigma}(\sigma)=0$ in Eq.~\plaineqref{eq:method:sde_sigma} yields the probability-flow ODE in
$\sigma$:
\begin{equation}
d\mathbf{x}
=
-\sigma\,\nabla_{\mathbf{x}}\log p(\mathbf{x};\sigma)\,d\sigma.
\label{eq:method:pf_sigma}
\end{equation}
Using the noise-prediction parameterization
$s_\theta(\mathbf{x},\sigma,c)=-\epsilon_\theta(\mathbf{x},\sigma,c)/\sigma$ to approximate the score gives
$d\mathbf{x}=\epsilon_{\theta}(\mathbf{x},\sigma,c)\,d\sigma$.
Applying the Euler--Maruyama discretization (Eq.~\plaineqref{eq:em_generic}) to Eq.~\plaineqref{eq:method:pf_sigma} gives
\begin{equation}
\mathbf{x}_{i+1}
=
\mathbf{x}_i
+
\Delta\sigma_i\,\epsilon_{\theta}(\mathbf{x}_i,\sigma_i,c),
\label{eq:method:eulerdiscrete_fp}
\end{equation}
where $\Delta\sigma_i\triangleq\sigma_{i+1}-\sigma_i<0$ is the step on the decreasing noise grid $\{\sigma_i\}$.
This matches standard Euler samplers under $\epsilon$-prediction \citep{karras2022elucidating}.

\paragraph{Quantization perturbation and its residual variability.}
With post-training quantization, the sampler uses $\hat{\epsilon}_{\theta}=\epsilon_{\theta}+\Delta\epsilon$.
Substituting this into Eq.~\plaineqref{eq:method:eulerdiscrete_fp} gives
\begin{equation}
\mathbf{x}_{i+1}
=
\mathbf{x}_i
+
\Delta\sigma_i\,\epsilon_{\theta}(\mathbf{x}_i,\sigma_i,c)
+
\Delta\sigma_i\,\Delta\epsilon_i,
\label{eq:method:eulerdiscrete_noise}
\end{equation}
where $\Delta\epsilon_i$ denotes the quantization noise term.
Let $V_{\sigma_i}$ denote the calibrated conditional residual-variance statistic defined in
Sec.~\ref{sec:method:stats}. To translate this statistic into a correction scale, we introduce an auxiliary
Gaussian perturbation $\boldsymbol{\eta}_i\sim\mathcal{N}(\mathbf{0},V_{\sigma_i}\mathbf{I})$, independent of
the current state and of the auxiliary perturbations at other steps. The increment
$\Delta\sigma_i\boldsymbol{\eta}_i$ has covariance $(\Delta\sigma_i)^2V_{\sigma_i}\mathbf{I}$.
We use this stochastic surrogate to design the correction.

\paragraph{Matching the auxiliary perturbation variance.}
Consider an Euler--Maruyama step of the auxiliary SDE in Eq.~\plaineqref{eq:method:sde_sigma}, whose
diffusion coefficient is $\sigma_i\sqrt{2\tilde{\beta}_{\sigma,i}}$:
\begin{equation}
\mathbf{x}_{i+1}
=
\mathbf{x}_i
+
(\cdots)\Delta\sigma_i
+
\sigma_i\sqrt{2\tilde{\beta}_{\sigma,i}}\;\sqrt{|\Delta\sigma_i|}\;\mathbf{z}_i,
\label{eq:method:em_sigma}
\end{equation}
where $\mathbf{z}_i\sim\mathcal{N}(\mathbf{0},\mathbf{I})$ and $(\cdots)$ collects the drift terms.
Matching the per-coordinate variance $2\tilde{\beta}_{\sigma,i}\sigma_i^2|\Delta\sigma_i|$ of its diffusion
increment to the auxiliary variance $(\Delta\sigma_i)^2V_{\sigma_i}$ yields
\begin{equation}
\tilde{\beta}_{\sigma,i}
=
\frac{|\Delta\sigma_i|}{2\sigma_i^2}\,V_{\sigma_i}.
\label{eq:method:beta_tilde}
\end{equation}

\paragraph{Correction scale from the paired drift.}
In Eq.~\plaineqref{eq:method:sde_sigma}, the matched diffusion strength also sets the paired
noise-decay drift. Both drift terms are proportional to the score, so their sum is
$-\sigma_i(1+\tilde{\beta}_{\sigma,i}\sigma_i)\nabla_{\mathbf{x}}\log p\,d\sigma$.
This pairing motivates rescaling the deterministic Euler update by $(1+c_i)$, with
\begin{equation}
c_i
\;\triangleq\;
\tilde{\beta}_{\sigma,i}\,\sigma_i
=
\frac{|\Delta\sigma_i|}{2\sigma_i}\,V_{\sigma_i}.
\label{eq:method:drift_scale}
\end{equation}

\paragraph{Final Q-Drift update.}
The final Q-Drift step is
\begin{equation}
\mathbf{x}_{i+1} = \mathbf{x}_i + \Delta\sigma_i\,(1+c_i)\,\hat{\epsilon}_{\theta}(\mathbf{x}_i,\sigma_i,c), \quad \text{with} \;\; c_i = \frac{|\Delta\sigma_i|}{2\sigma_i}\,V_{\sigma_i}.
\label{eq:method:qdrift_eulerdiscrete}
\end{equation}
Q-Drift preserves the direction of the quantized Euler update while adjusting its magnitude.
The correction increases with the calibrated residual variance and the relative step size
$|\Delta\sigma_i|/\sigma_i$, and vanishes when $V_{\sigma_i}=0$.
Q-Drift applies the resulting multiplier to the quantized prediction, requiring only the precomputed
$V_{\sigma_i}$ and no additional noise injection.

\paragraph{Extension to various samplers.}
We adapt this rescaling principle to flow matching and DPM-Solver++, as detailed in
Appendix~\ref{sec:app:flowmatching} and Appendix~\ref{sec:app:dpm_solver_qdrift}, respectively.

\subsection{Calibration: Estimating the Conditional Variance Statistic}
\label{sec:method:stats}

Offline, on a calibration set, we run a full-precision sampling trajectory and record at every step $\sigma_i$ the
paired outputs $\epsilon_{\theta}(\mathbf{x},\sigma_i,c)$ and $\hat{\epsilon}_{\theta}(\mathbf{x},\sigma_i,c)$,
whose difference is the quantization noise $\Delta\epsilon_i$.
The statistic $V_{\sigma_i}$ measures the residual variability of quantization error after conditioning
on the quantized output:
\begin{equation}
V_{\sigma_i}
\;\triangleq\;
\mathbb{E}\Bigl[\mathrm{Var}\bigl(\Delta\epsilon_i \mid \hat{\epsilon}_{\theta}(\mathbf{x},\sigma_i,c)\bigr)\Bigr].
\label{eq:method:V_def}
\end{equation}

Here, variance denotes the average per-coordinate conditional variance under the covariance model in
Sec.~\ref{sec:prelim:joint-simplify}, and the expectation is over the calibration distribution.
The conditional mean minimizes squared prediction error given the quantized output; its residual variance
quantifies the variability remaining after that prediction. We estimate this statistic using the fitted joint
Gaussian model in Eq.~\plaineqref{eq:cond_cov_d2}.

Q-Drift uses this variance statistic without subtracting the conditional mean, yielding one correction
factor per step. We compare this approach with prior corrections in Sec.~\ref{sec:exp:sampler_side} and examine
the choice of variance statistic in Sec.~\ref{sec:exp:calibration_ablation}.

\subsection{Algorithm}
\label{sec:method:algo}
Algorithm~\ref{alg:qdrift} in Appendix~\ref{sec:app:algo} summarizes the overall Q-Drift procedure: an offline pass that records paired full-precision and quantized outputs along a sampling trajectory and fits $V_{\sigma_i}$ per step, and an online pass that applies Eq.~\plaineqref{eq:method:qdrift_eulerdiscrete}.

\section{Experiments}
\label{sec:exp}

\subsection{Experimental Setup}
\label{sec:exp:setup}

\paragraph{Models and samplers.}
Following SVDQuant~\citep{li2025svdquant}, we evaluate Q-Drift on six text-to-image models spanning DiT and U-Net backbones:
FLUX.1-dev (20 steps), its distilled variant FLUX.1-schnell (4 steps), PixArt-$\Sigma$~\citep{chen2024pixartsigma} (20 steps) and Sana~\citep{xie2024sana} (20 steps) for DiT, and SDXL~\citep{podell2024sdxl} (30 steps) with its distilled variant SDXL-Turbo~\citep{sauer2024add} (4 steps) for U-Net.
Each model keeps its default inference configuration and its standard sampler: flow matching for the two FLUX models, first-order Euler for SDXL and SDXL-Turbo, and DPM-Solver++(2M, midpoint) for PixArt-$\Sigma$ and Sana, following their official pipelines~\citep{lu2022dpm_solver_plus_plus}.

\paragraph{Datasets.}
Following the SVDQuant protocol~\citep{li2025svdquant}, we sample 5K prompts for evaluation and a disjoint set of 1K prompts for calibration from MJHQ-30K~\citep{mjhq30k}.
The two subsets use different random seeds, and all methods are evaluated on the same evaluation subset for fair comparison.

\paragraph{Baselines and compared variants.}
Our quantized baselines are SVDQuant~\citep{li2025svdquant} and MixDQ~\citep{zhao2024mixdq}, representative recent PTQ methods for text-to-image diffusion with publicly available implementations.
For each model we report FP (full precision), Quantized (the quantized model named in the model column), and Q-Drift (ours, applied on top of that same quantized model at sampling time).
Q-Drift involves no fine-tuning and does not modify network weights; it only corrects the sampler dynamics.

\paragraph{Metrics.}
We measure both image fidelity and prompt alignment.
For fidelity, we report FID (lower is better) \citep{heusel2017fid}, computed on images generated from the MJHQ prompt set.
For prompt alignment, we report CLIP Score (higher is better) \citep{hessel2021clipscore,radford2021clip}.
We additionally report PSNR, LPIPS, and SSIM for similarity between each method's outputs and the FP outputs.

\begin{table}[!t]
  \caption{Main results on MJHQ-30K under aggressive quantization settings.
  SVDQuant rows use W3A4 quantization, and the MixDQ row uses W4A8 quantization.
  Boldface in the FID column indicates the lower value between the quantized baseline and Q-Drift for each setting.}
  \label{tab:exp:main}
  \centering

  \begin{tabular}{llccccc}
    \toprule
    Model & Method & FID$\downarrow$ & CLIP$\uparrow$ & PSNR$\uparrow$ & LPIPS$\downarrow$ & SSIM$\uparrow$ \\
      \midrule
	    FLUX.1-dev & FP & 20.68 & 25.80 & -- & -- & -- \\
	    (SVDQuant W3A4) & Quantized & 24.14 & 24.66 & 16.13 & 0.491 & 0.621 \\
	       & Q-Drift & \textbf{24.06} & 24.69 & 16.15 & 0.488 & 0.623 \\
	    \midrule
	    FLUX.1-schnell & FP & 19.18 & 26.55 & -- & -- & -- \\
	    (SVDQuant W3A4) & Quantized & 23.10 & 25.43 & 13.75 & 0.529 & 0.519 \\
	       & Q-Drift & \textbf{22.13} & 25.57 & 13.75 & 0.520 & 0.528 \\
	    \midrule
		  SDXL & FP & 17.20 & 27.40 & -- & -- & -- \\
		  (SVDQuant W3A4) & Quantized & 31.73 & 26.39 & 14.88 & 0.549 & 0.552 \\
		     & Q-Drift & \textbf{30.69} & 26.40 & 14.88 & 0.547 & 0.551 \\
	    \midrule
		  SDXL-Turbo & FP & 24.77 & 26.39 & -- & -- & -- \\
		  (SVDQuant W3A4) & Quantized & 29.17 & 26.38 & 14.30 & 0.455 & 0.456 \\
		     & Q-Drift & \textbf{27.64} & 26.38 & 14.35 & 0.457 & 0.457 \\
	    \midrule
		  SDXL-Turbo & FP & 24.77 & 26.39 & -- & -- & -- \\
		  (MixDQ W4A8) & Quantized & 28.37 & 25.90 & 15.68 & 0.402 & 0.516 \\
		     & Q-Drift & \textbf{27.38} & 25.87 & 12.62 & 0.587 & 0.356 \\
	    \midrule
	    PixArt-$\Sigma$ & FP & 16.52 & 26.71 & -- & -- & -- \\
	    (SVDQuant W3A4) & Quantized & 48.85 & 24.41 & 13.53 & 0.547 & 0.503 \\
	       & Q-Drift & \textbf{44.06} & 24.56 & 13.61 & 0.544 & 0.504 \\
	    \midrule
	    Sana & FP & 15.98 & 27.30 & -- & -- & -- \\
	    (SVDQuant W3A4) & Quantized & 17.32 & 27.19 & 16.71 & 0.284 & 0.616 \\
	       & Q-Drift & \textbf{16.54} & 27.20 & 16.69 & 0.286 & 0.612 \\
		  \bottomrule
		  \end{tabular}

\end{table}

\subsection{Main Results}
\label{sec:exp:main}

Table~\ref{tab:exp:main} summarizes results across model families and quantization settings.
Q-Drift improves FID over the corresponding quantized baseline in all seven settings of Table~\ref{tab:exp:main}, across the evaluated diffusion backbones, PTQ methods, and samplers, and the effect is clearest when read as recovery from quantization-induced degradation.
For SDXL-Turbo under SVDQuant W3A4 it reduces FID from 29.17 to 27.64 against a full-precision reference of
24.77; on PixArt-$\Sigma$, where aggressive quantization degrades quality the most, it gives the largest absolute
gain, from 48.85 to 44.06; and even in a comparatively robust setting such as Sana it improves FID from 17.32 to
16.54, recovering roughly 58\% of the gap to full precision.
Additional results under milder quantization settings are provided in Appendix~\ref{sec:app:main_additional}.
CLIP scores remain essentially unchanged, suggesting that Q-Drift preserves prompt alignment while improving
distributional fidelity, and since it applies only a lightweight step-wise drift rescaling, it incurs negligible
inference overhead.
As reported in Appendix~\ref{sec:app:kid_ci}, Q-Drift also lowers KID in all seven settings of
Table~\ref{tab:exp:main}. Paired $95\%$ bootstrap intervals for
$\Delta\mathrm{FID}=\mathrm{FID}(\text{Q-Drift})-\mathrm{FID}(\text{Quantized})$ lie entirely below zero in
six settings; the exception is FLUX.1-dev, whose interval includes zero.
Q-Drift is designed with marginal preservation as its objective, and we evaluate its effect on
distributional fidelity using FID and KID. Similarity to individual FP outputs, measured by PSNR, LPIPS, and SSIM,
is not the design objective.

\begin{figure}[!t]
  \centering

  \newcommand{\visw}{\dimexpr(\linewidth-8\tabcolsep)/5\relax}
  \begin{tabular}{@{}ccccc@{}}
		\includegraphics[width=\visw]{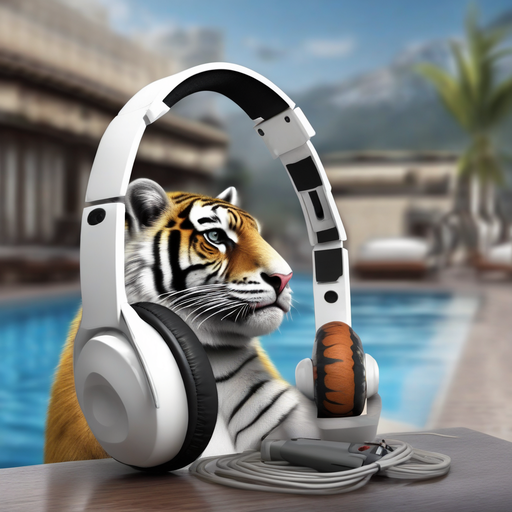} &
		\includegraphics[width=\visw]{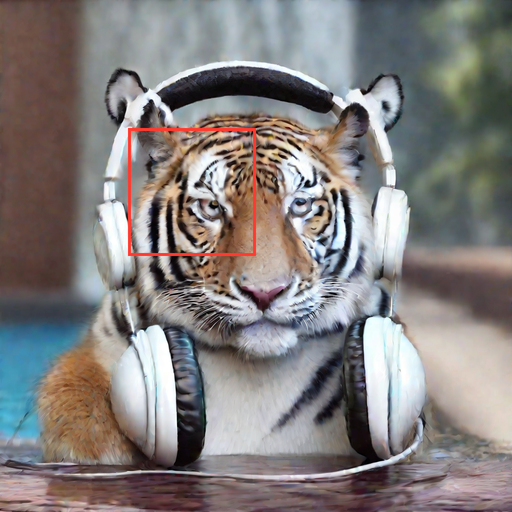} &
		\includegraphics[width=\visw]{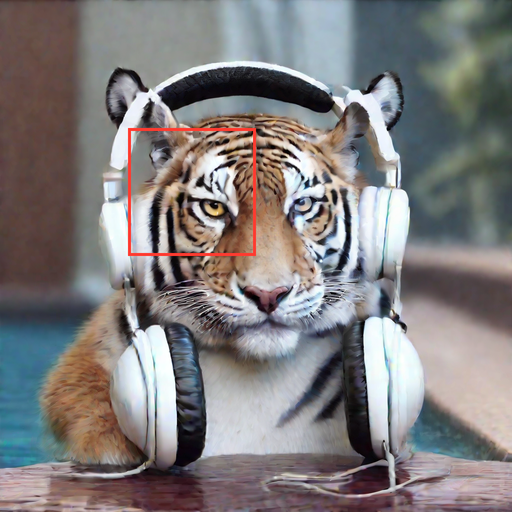} &
    \includegraphics[width=\visw]{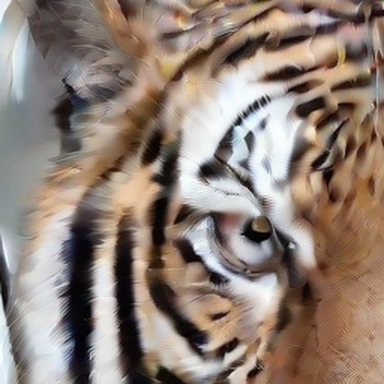} &
    \includegraphics[width=\visw]{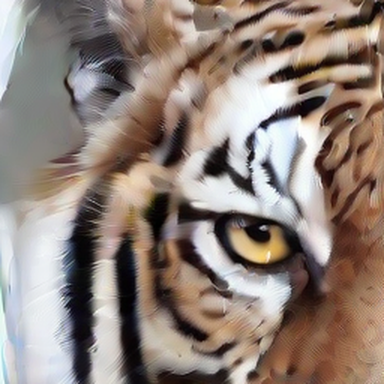} \\
				\addlinespace[1pt]
		\includegraphics[width=\visw]{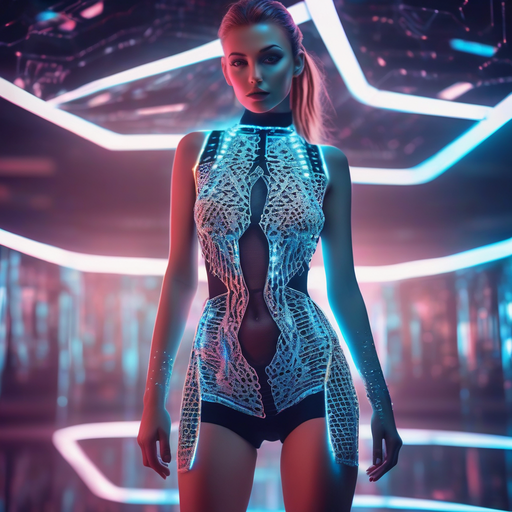} &
		\includegraphics[width=\visw]{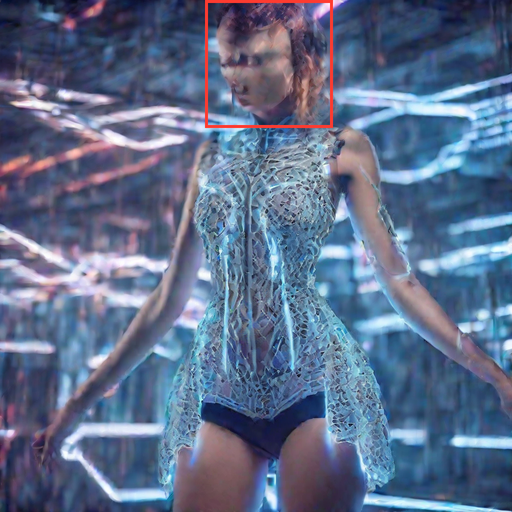} &
		\includegraphics[width=\visw]{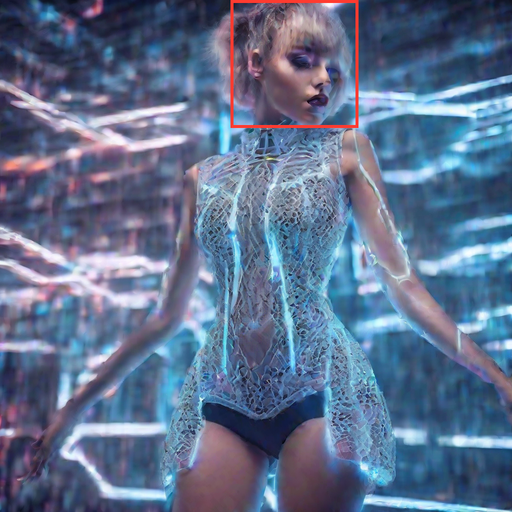} &
    \includegraphics[width=\visw]{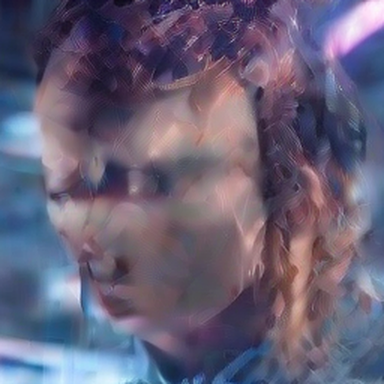} &
    \includegraphics[width=\visw]{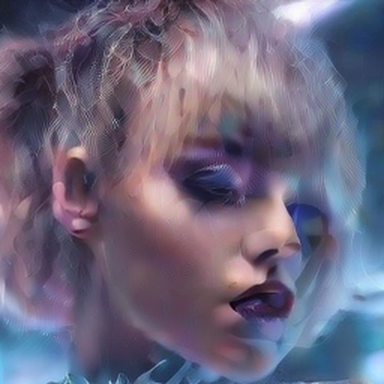} \\
				\addlinespace[1pt]
		\includegraphics[width=\visw]{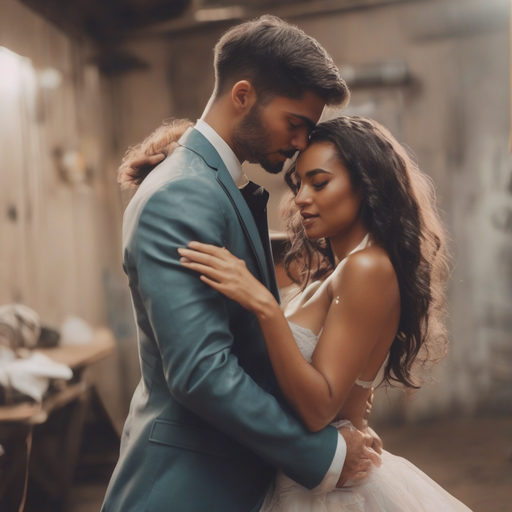} &
		\includegraphics[width=\visw]{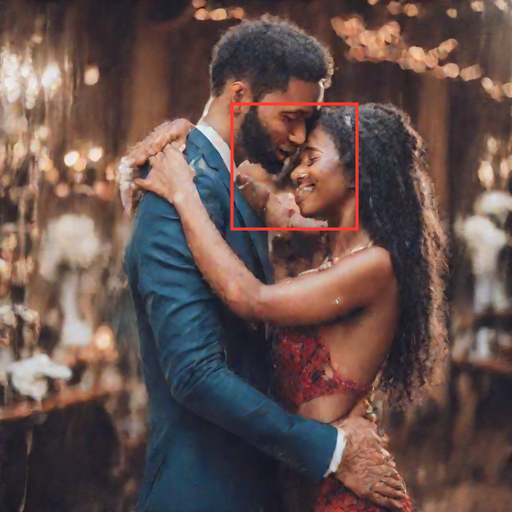} &
		\includegraphics[width=\visw]{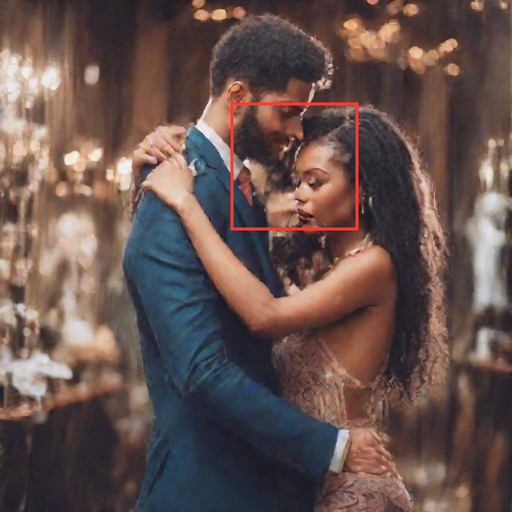} &
    \includegraphics[width=\visw]{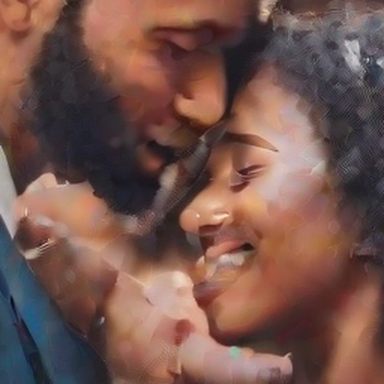} &
    \includegraphics[width=\visw]{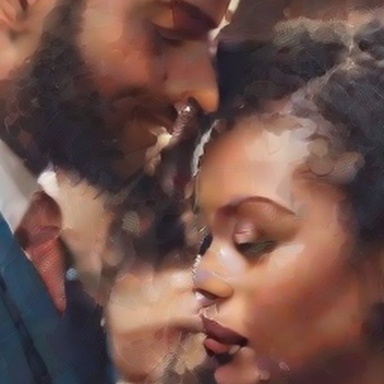} \\
				\addlinespace[1pt]
		\includegraphics[width=\visw]{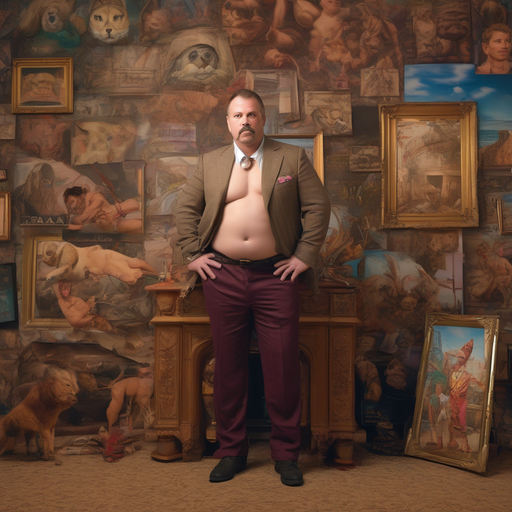} &
		\includegraphics[width=\visw]{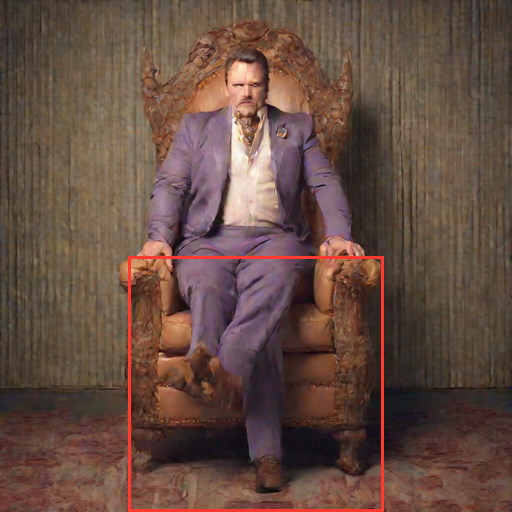} &
		\includegraphics[width=\visw]{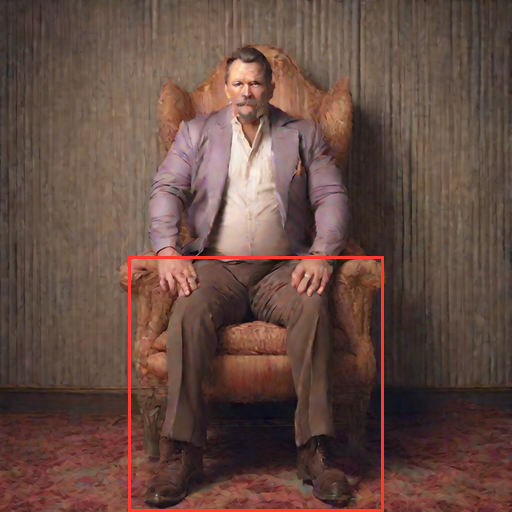} &
    \includegraphics[width=\visw]{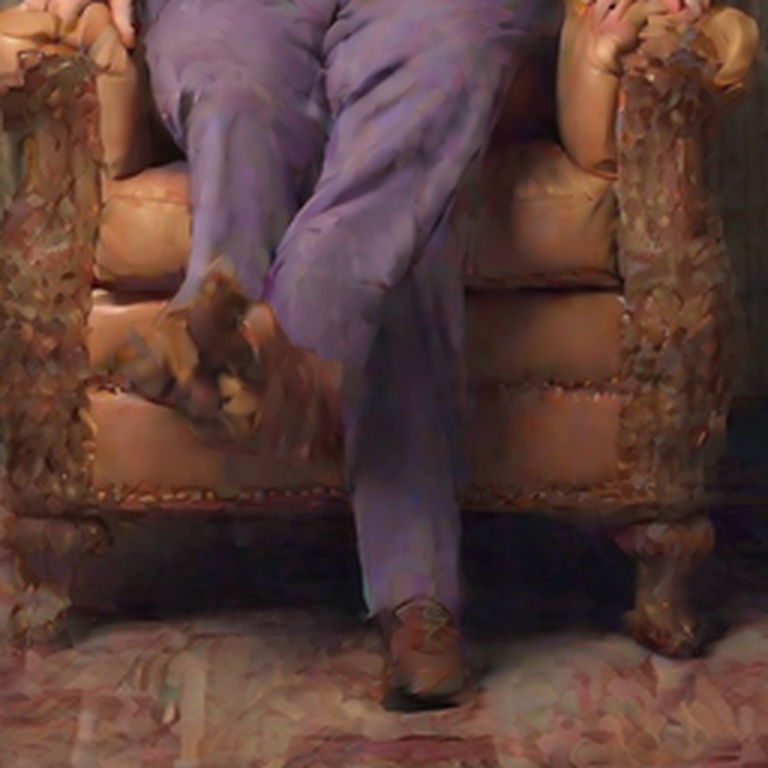} &
    \includegraphics[width=\visw]{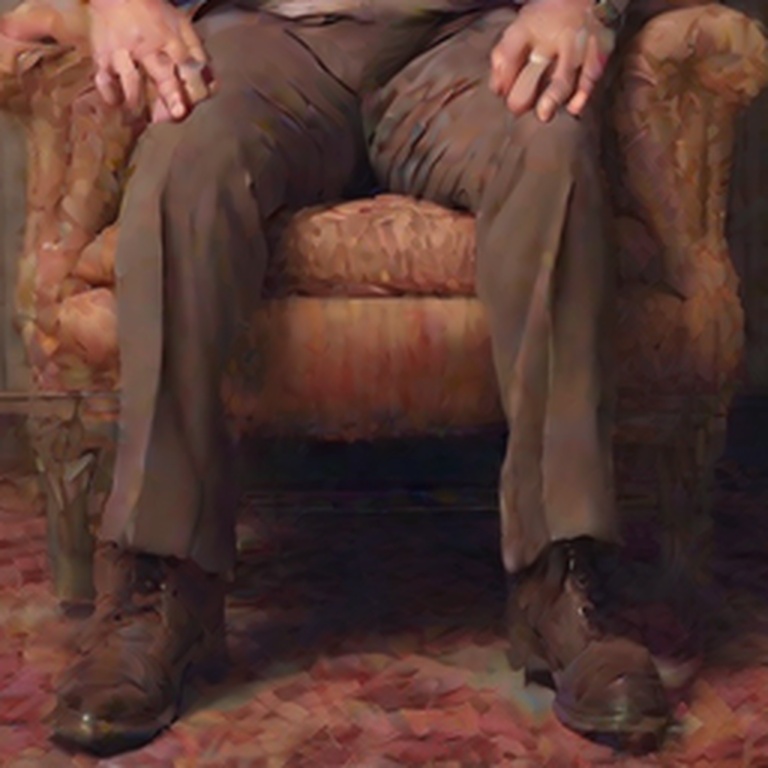} \\
				\addlinespace[1pt]
		\includegraphics[width=\visw]{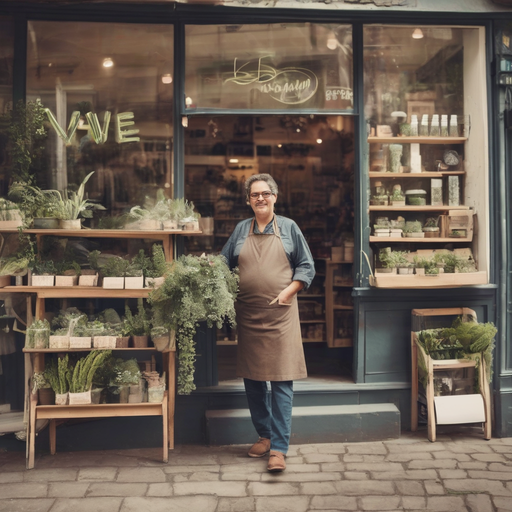} &
		\includegraphics[width=\visw]{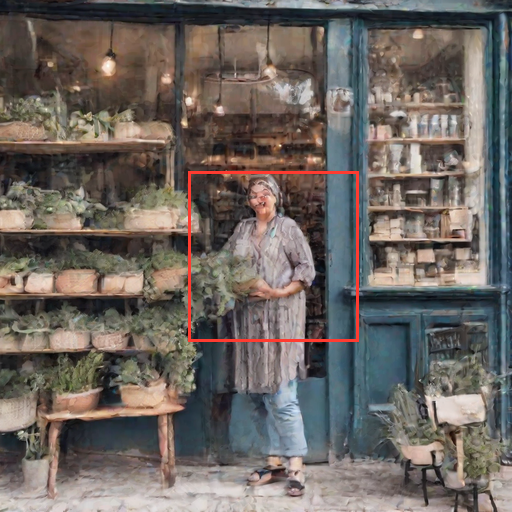} &
		\includegraphics[width=\visw]{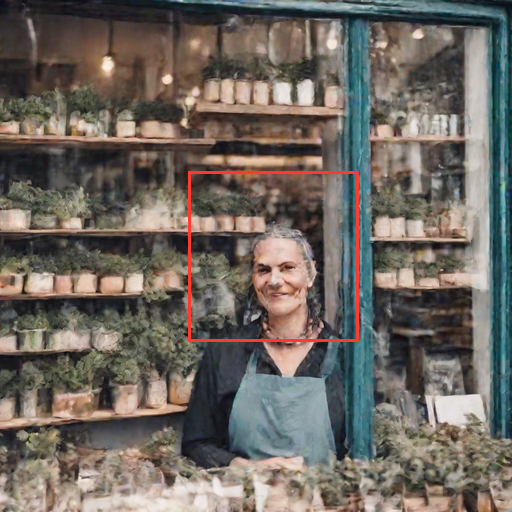} &
    \includegraphics[width=\visw]{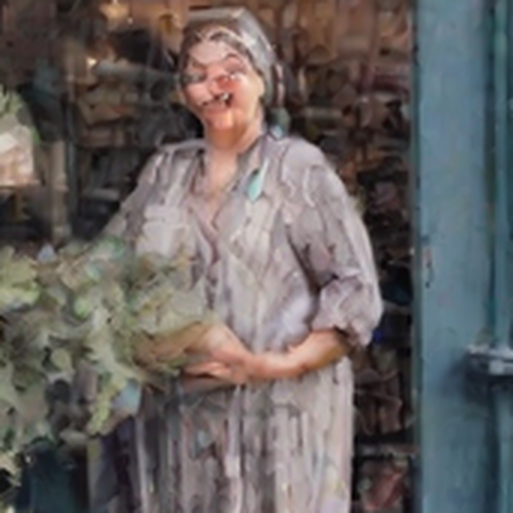} &
    \includegraphics[width=\visw]{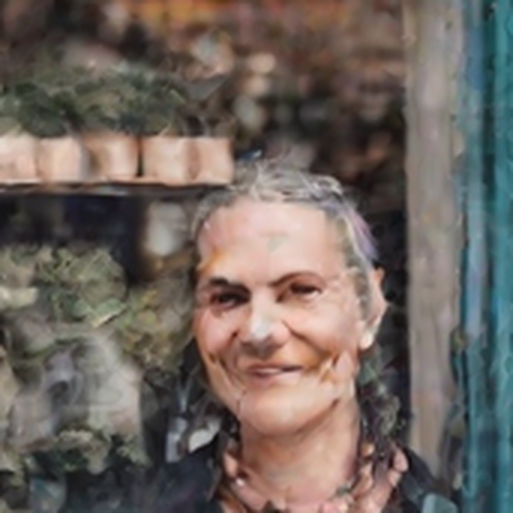} \\
      {FP} & {Quantized} & {Q-Drift} & \shortstack{Quantized\\zoom} & \shortstack{Q-Drift\\zoom} \\
	  \end{tabular}
	  \caption{\textbf{Visual comparisons on SDXL (SVDQuant W3A4).}
    For readability, prompt texts are provided in Appendix~\ref{sec:app:visual_prompts}. Red boxes indicate the regions shown in the zoomed panels.}
	  \label{fig:exp:visual_sdxl_w3a4}
\end{figure}

\subsection{Qualitative Visual Comparisons}
\label{sec:app:visual_comparisons}

Fig.~\ref{fig:exp:visual_sdxl_w3a4} compares SDXL (SVDQuant W3A4) outputs under identical prompts and initial noise.
Q-Drift yields subtle but consistent improvements relative to the quantized baseline: it sharpens local detail in the first three rows and resolves an anatomically implausible leg overlap in the fourth.
In the fifth row it produces a structure that differs noticeably from both FP16 and the quantized sample while preserving prompt alignment and perceptual quality, a reminder that the correction reshapes the trajectory
rather than tracking the FP sample.

\begin{table}[!t]
  \centering
  \caption{\textbf{Comparison with prior sampler-side corrections} on SDXL (SVDQuant W3A4, Euler).}
  \label{tab:exp:sampler_side}

  \begin{tabular}{lcc}
    \toprule
    Sampler-side method & FID$\downarrow$ & CLIP$\uparrow$ \\
    \midrule
    Quantized baseline & 31.73 & 26.39 \\
    PTQD & 31.51 & 26.35 \\
    QNCD & 32.01 & 26.37 \\
    D$^2$-DPM, deterministic & 74.28 & 25.65 \\
    D$^2$-DPM, stochastic & 82.87 & 25.32 \\
    \midrule
    \textbf{Q-Drift} & \textbf{30.69} & 26.40 \\
    \bottomrule
  \end{tabular}

\end{table}

\subsection{Comparison with Prior Sampler-Side Correction Methods}
\label{sec:exp:sampler_side}

\paragraph{Empirical comparison.}
Table~\ref{tab:exp:sampler_side} compares prior sampler-side corrections with Q-Drift on SDXL
(SVDQuant W3A4, Euler), adapting each method to this sampler as faithfully as we can.
We found no public implementation of TAC-Diffusion, and for QNCD we omit the intra-step feature smoothing, since
SVDQuant already applies it.
Q-Drift gives the lowest FID, at a CLIP score comparable to the quantized baseline, while every prior correction
lands above the quantized baseline it was meant to improve, except PTQD, which is close to it.
We next examine the correction mechanisms and their behavior in this setting.

\paragraph{Prediction variance under conditional-mean correction.}
Prior corrections aim to predict and remove the quantization error through different mechanisms
(Table~\ref{tab:exp:positioning}, Appendix~\ref{sec:app:mechanisms}).
Under squared error, the optimal correction based on the quantized output is the conditional mean
$\mathbb{E}[\Delta\epsilon\mid\hat{\epsilon}]$.
Let $\tilde{\epsilon}=\hat{\epsilon}-\mathbb{E}[\Delta\epsilon\mid\hat{\epsilon}]
=\mathbb{E}[\epsilon_{\theta}\mid\hat{\epsilon}]$. The law of total variance gives
\begin{equation}
\mathrm{Var}(\epsilon_{\theta})
=\mathrm{Var}(\tilde{\epsilon})+\mathbb{E}\bigl[\mathrm{Var}(\Delta\epsilon\mid\hat{\epsilon})\bigr].
\label{eq:exp:total_variance}
\end{equation}
Thus, conditional-mean correction reduces prediction variance relative to the full-precision output.

D$^2$-DPM addresses residual variability in two ways. Its stochastic variant subtracts a random sample from
$\mathcal{N}(\mu_{\mathrm{cond}},\Sigma_{\mathrm{cond}})$ rather than $\mu_{\mathrm{cond}}$ itself, restoring
prediction variance under the conditional Gaussian model. Its deterministic variant subtracts
$\mu_{\mathrm{cond}}$ and adjusts the sampler's diffusion term, as PTQD's variance schedule calibration does
for the injected noise. Such diffusion adjustments are unavailable in the deterministic Euler, flow-matching,
and DPM-Solver++ samplers used here, whereas stochastic error subtraction introduces randomness into the update.

D$^2$-DPM yields $74.28$ FID in its deterministic variant and $82.87$ in its stochastic variant.
In the stochastic variant, the injected noise amounts to $10.5\%$--$54.0\%$ of the update magnitude across steps.
Neither variant improves FID in this setting.
PTQD remains close to the baseline.
Its scaling mechanism is switched off here by its own rule that resets a negative coefficient to zero, so all that
remains is a fixed per-channel unconditional mean, and subtracting a constant shifts the prediction without
shrinking its variance.

Calibration-based error predictors must also generalize across conditioning inputs. Their effectiveness depends
on the calibration distribution and the information used to predict the error.
TAC-Diffusion~\citep{yao2024tac} specifically reports that prompt-conditioned quantization noise is too diverse
to cover with a calibration set of practical size, and responds by correcting only the unconditional branch
of the prediction.

Q-Drift instead uses residual variance to rescale the deterministic update, without subtracting a predicted error or injecting noise.

\begin{table}[!t]
  \centering
  \caption{\textbf{Calibration size and design ablations} on SDXL (SVDQuant W3A4).}
  \label{tab:exp:calib_K}

  \begin{tabular}{lcc}
    \toprule
    Calibration / variant & FID$\downarrow$ & CLIP$\uparrow$ \\
    \midrule
    Quantized baseline (no correction) & 31.73 & 26.39 \\
    \midrule
    Q-Drift, 1K-prompt reference & 30.69 & 26.40 \\
    \midrule
    $K=10$, max $\sum_i |\Delta c_i|$ & 30.85 & 26.36 \\
    $K=10$, min $\sum_i \Delta c_i$ & 30.67 & 26.36 \\
    $K=10$, max $\sum_i \Delta c_i$ & 30.52 & 26.37 \\
    \midrule
    scalar unconditional variance & 31.95 & 26.35 \\
    channel-wise conditional variance & 30.82 & 26.36 \\
    \bottomrule
  \end{tabular}

\end{table}

\subsection{Calibration Efficiency and Ablation Studies}
\label{sec:exp:calibration_ablation}

\paragraph{Calibration Efficiency.}
Q-Drift calibrates only the per-step scalar factors $c_i$, whereas prior methods fit sample-level error corrections.
In our comparison, PTQD and D$^2$-DPM use $1{,}000$ paired runs; TAC-Diffusion reports using $256$ samples for text-to-image generation.
We test smaller calibration sets on SDXL (SVDQuant W3A4), re-estimating $\{c_i\}$ from each of 200 subsets
of $K=10$ paired runs drawn from the same 1K-pair calibration pool. Taking the 1K-prompt factors $\{c_i^{(\mathrm{ref})}\}$ as reference, we define
$\Delta c_i\triangleq c_i-c_i^{(\mathrm{ref})}$ and select three stress cases: the subset maximizing
$\sum_i|\Delta c_i|$ and those maximizing and minimizing $\sum_i\Delta c_i$.
The $K=10$ rows in Table~\ref{tab:exp:calib_K} report these stress cases rather than typical subsets.
All three yield FID within $0.17$ of the reference, with similar CLIP scores, supporting the use of $K=10$
in this setting.

\paragraph{Ablation Study.}
Both design ablations use the same 1K-prompt calibration set as the scalar conditional-variance reference.
Unconditional variance includes variability of the conditional mean, increasing the correction scale.
Under the same calibration and evaluation setup, it yields $31.95$ FID versus $30.69$ with conditional variance
(Table~\ref{tab:exp:calib_K}), supporting the conditional statistic here.
The channel-wise variant estimates conditional variance separately for each latent channel and applies
the corresponding correction channel-wise, yielding $30.82$ FID versus $30.69$ for the scalar form.
The finer granularity provides no FID gain here, supporting one factor per step.

\section{Conclusion}
We presented Q-Drift, a quantization-aware drift correction that revisits quantized diffusion sampling from a distributional perspective.
Across the models, samplers, and PTQ methods, Q-Drift improves FID at comparable
CLIP scores, with negligible inference overhead. Our SDXL study further demonstrates effective calibration with
as few as 10 paired runs.
Relaxing its calibration-friendly assumptions, such as the simplified covariance structure, is left to future work.

\bibliographystyle{iclr2027_conference}
\bibliography{reference}

\clearpage
\appendix
\section*{Appendix}

\section{Q-Drift Algorithm}
\label{sec:app:algo}

Algorithm~\ref{alg:qdrift} summarizes Q-Drift's offline calibration and online sampling procedure for a first-order Euler sampler.

\begin{algorithm}
  \caption{Q-Drift: Quantization-Aware Drift Correction (Calibration + Sampling)}
  \label{alg:qdrift}
  \begin{algorithmic}[1]
    \STATE {\bfseries Input:} full-precision denoiser $\epsilon_{\theta}$, quantized denoiser $\hat{\epsilon}_{\theta}$,
    conditioning $c$, schedule $\{\sigma_i\}_{i=0}^{M}$, calibration set $\mathcal{D}_{\mathrm{cal}}$, number of samples $N$.
    \STATE {\bfseries Output:} $\{\mathbf{x}^{(n)}_{\sigma_M}\}_{n=1}^{N}$.

    \STATE {\bfseries Offline calibration.}
    \FOR{each conditioning $c_{\mathrm{cal}}\in\mathcal{D}_{\mathrm{cal}}$}
      \STATE Sample $\mathbf{x}_{\sigma_0}\sim \mathcal{N}(\mathbf{0}, \sigma_0^2\mathbf{I})$.
      \FOR{$i=0$ {\bfseries to} $M-1$}
        \STATE $\epsilon \leftarrow \epsilon_{\theta}(\mathbf{x}_{\sigma_i},\sigma_i,c_{\mathrm{cal}})$
        \STATE $\hat{\epsilon} \leftarrow \hat{\epsilon}_{\theta}(\mathbf{x}_{\sigma_i},\sigma_i,c_{\mathrm{cal}})$
        \STATE $\Delta\epsilon \leftarrow \hat{\epsilon}-\epsilon$
        \STATE Store paired outputs $(\hat{\epsilon},\Delta\epsilon)$ at $\sigma_i$
        \STATE $\Delta\sigma_i \leftarrow \sigma_{i+1}-\sigma_i$
        \STATE $\mathbf{x}_{\sigma_{i+1}} \leftarrow \mathbf{x}_{\sigma_i} + \Delta\sigma_i\,\epsilon$
      \ENDFOR
    \ENDFOR
    \FOR{$i=0$ {\bfseries to} $M-1$}
      \STATE Fit the Gaussian model to stored pairs at $\sigma_i$ (Sec.~\ref{sec:prelim:joint-gauss}--\ref{sec:prelim:joint-simplify})
      \STATE Compute $V_{\sigma_i}$ (Eq.~\plaineqref{eq:method:V_def})
    \ENDFOR

    \STATE {\bfseries Online sampling (first-order Euler sampler).}
    \FOR{$n=1$ {\bfseries to} $N$}
      \STATE Sample $\mathbf{x}^{(n)}_{\sigma_0}\sim \mathcal{N}(\mathbf{0}, \sigma_0^2\mathbf{I})$.
      \FOR{$i=0$ {\bfseries to} $M-1$}
        \STATE $\Delta\sigma_i \leftarrow \sigma_{i+1}-\sigma_i$
        \STATE $\hat{\epsilon} \leftarrow \hat{\epsilon}_{\theta}(\mathbf{x}^{(n)}_{\sigma_i},\sigma_i,c)$
        \STATE $c_i \leftarrow \dfrac{|\Delta\sigma_i|}{2\sigma_i}\,V_{\sigma_i}$
        \STATE $\mathbf{x}^{(n)}_{\sigma_{i+1}} \leftarrow \mathbf{x}^{(n)}_{\sigma_i} + \Delta\sigma_i\,(1+c_i)\,\hat{\epsilon}$
      \ENDFOR
    \ENDFOR
    \STATE {\bfseries return} $\{\mathbf{x}^{(n)}_{\sigma_M}\}_{n=1}^{N}$.
  \end{algorithmic}
\end{algorithm}

\begin{table}[!t]
  \centering
  \caption{\textbf{Mechanisms of sampler-side corrections for quantized diffusion.}
  Methods that combine several mechanisms are listed one row per mechanism.
  QNCD's intra-step feature smoothing is omitted, as it acts inside the network rather than on the
  sampler.
  ``stochastic'' marks mechanisms that require $\sigma_t>0$; ``studied'' refers to the deterministic Euler,
  flow-matching, and DPM-Solver++ samplers evaluated here.}
  \label{tab:exp:positioning}

  \begin{tabular}{llp{0.28\linewidth}c}
    \toprule
    Method (mechanism) & Statistic estimated & How it enters the sampler & Samplers \\
    \midrule
    PTQD (CNC) & correlation coeff.\ $k$ & divides the output by $1{+}k$ & any \\
    PTQD (BC) & channel-wise mean of $\Delta\epsilon$ & subtracted from the output & any \\
    PTQD (VSC) & residual variance $\sigma_q^2$ & shrinks the injected noise $\sigma_t$ & stochastic \\
    \addlinespace[1pt]
    TAC-Diff.\ (NER) & channel-wise coeff.\ $K_t$ & rescales the output & any \\
    TAC-Diff.\ (IBC) & input bias & corrects the sampler input $x_t$ & any \\
    \addlinespace[1pt]
    QNCD (inter) & per-sample noise estimate & subtracted from the output & any \\
    \addlinespace[1pt]
    D$^2$-DPM (TSQNM) & conditional mean $\mu_{\mathrm{cond}}$ & subtracted from the output & any \\
    D$^2$-DPM (S-D$^2$) & conditional cov.\ $\Sigma_{\mathrm{cond}}$ & sampled, then subtracted & any \\
    D$^2$-DPM (D-D$^2$) & conditional cov.\ $\Sigma_{\mathrm{cond}}$ & shrinks the diffusion term $\Sigma_t$ & stochastic \\
    \midrule
    \textbf{Q-Drift} & residual variance $V_{\sigma_i}$ & rescales the deterministic update & studied \\
    \bottomrule
  \end{tabular}

\end{table}

\section{Mechanisms of Prior Sampler-Side Corrections}
\label{sec:app:mechanisms}

Table~\ref{tab:exp:positioning} summarizes the statistics and correction mechanisms used by prior sampler-side methods and Q-Drift.

\section{Additional Derivations and Extensions}
\label{sec:app:derivations}

\subsection{Extension to Flow-Matching Sampler}
\label{sec:app:flowmatching}

Many flow-matching and rectified-flow samplers use a first-order update of the form
\begin{equation}
\mathbf{x}_{i+1}
=
\mathbf{x}_i
+
\Delta\sigma_i\,v_{\theta}(\mathbf{x}_i,\sigma_i,c),
\label{eq:app:flowmatching:euler_update}
\end{equation}
where $\sigma$ denotes the sampler's interpolation coordinate and $v_{\theta}$ is its velocity field.
The update has the same numerical form as the Euler update in Eq.~\plaineqref{eq:method:eulerdiscrete_fp}.

\paragraph{Q-Drift for Flow-matching.}
With quantization, the sampler uses $\hat{v}_{\theta}=v_{\theta}+\Delta v$.
We adapt the Euler rescaling rule by using the conditional residual variance of $\Delta v_i$ given
$\hat{v}_{\theta}$ as $V_{\sigma_i}$ in Eq.~\plaineqref{eq:method:V_def}.
With $c_i=\frac{|\Delta\sigma_i|}{2\sigma_i}V_{\sigma_i}$, the update is
\begin{equation}
\mathbf{x}_{i+1}
=
\mathbf{x}_i
+
\Delta\sigma_i\,(1+c_i)\,\hat{v}_{\theta}(\mathbf{x}_i,\sigma_i,c).
\label{eq:app:flowmatching:qdrift}
\end{equation}

\paragraph{Scope of the extension.}
A general flow-matching velocity does not satisfy the score--velocity relation used in the VE derivation.
We therefore treat this rescaling as an empirical extension to the flow-matching sampler.

\subsection{Q-Drift for DPM-Solver++ in Log-SNR Coordinates}
\label{sec:app:dpm_solver_qdrift}

We adapt Q-Drift to DPM-Solver++~\citep{lu2022dpm_solver_plus_plus} by propagating calibrated residual-variance
statistics through the solver's model-output coefficients and rescaling the baseline update in normalized
coordinates. The generalized SDE in log-SNR coordinates motivates the correction scale.

\paragraph{Log-SNR notation and VE reparameterization.}
We write the forward marginal as
\begin{equation}
\mathbf{x}_t=\alpha_t\mathbf{x}_0+\sigma_t\boldsymbol{\epsilon},
\qquad \boldsymbol{\epsilon}\sim\mathcal{N}(\mathbf{0},\mathbf{I}),
\label{eq:app:dpm:x_marginal}
\end{equation}
and define the log-SNR coordinate
\begin{equation}
\lambda_t\triangleq\log\frac{\alpha_t}{\sigma_t},
\qquad \frac{\sigma_t}{\alpha_t}=e^{-\lambda_t}.
\label{eq:app:dpm:logsnr}
\end{equation}
In normalized coordinates $\mathbf{y}_t\triangleq\mathbf{x}_t/\alpha_t$, this becomes
\begin{equation}
\mathbf{y}_t=\mathbf{x}_0+\bar{\sigma}_t\boldsymbol{\epsilon},
\qquad \bar{\sigma}_t\triangleq\frac{\sigma_t}{\alpha_t}=e^{-\lambda_t}.
\label{eq:app:dpm:y_marginal}
\end{equation}
The normalized forward marginals thus follow a VE-type path with unit signal scale and noise level
$\bar{\sigma}$~\citep{song2021score}. Sampling proceeds toward increasing $\lambda$.

\paragraph{Generalized SDEs in log-SNR coordinates.}
Applying Eq.~\plaineqref{eq:method:sde_sigma} to this normalized marginal path gives
\begin{equation}
\begin{aligned}
d\mathbf{y}
={}&-\bar{\sigma}\bigl(1+\tilde{\beta}_{\bar{\sigma}}\bar{\sigma}\bigr)
\nabla_{\mathbf{y}}\log p(\mathbf{y};\bar{\sigma})\,d\bar{\sigma}
+\bar{\sigma}\sqrt{2\tilde{\beta}_{\bar{\sigma}}}\,d\mathbf{w}_{\bar{\sigma}},
\end{aligned}
\label{eq:app:dpm:sde_barsigma}
\end{equation}
where $\mathbf{w}_{\bar{\sigma}}$ is a reverse-time Wiener process with increment covariance
$|d\bar{\sigma}|\mathbf{I}$. Using $d\bar{\sigma}=-\bar{\sigma}\,d\lambda$, define
$\tilde{\beta}_{\lambda}\triangleq\bar{\sigma}\tilde{\beta}_{\bar{\sigma}}\geq0$ and the exact score-based
quantity $\epsilon_*(\mathbf{y},\lambda)\triangleq-\bar{\sigma}\nabla_{\mathbf{y}}\log p(\mathbf{y};\bar{\sigma})$.
Then
\begin{equation}
d\mathbf{y}=-(1+\tilde{\beta}_{\lambda})\bar{\sigma}\epsilon_*(\mathbf{y},\lambda)\,d\lambda
+\bar{\sigma}\sqrt{2\tilde{\beta}_{\lambda}}\,d\mathbf{w}_{\lambda},
\label{eq:app:dpm:sde_lambda}
\end{equation}
where $\mathbf{w}_{\lambda}$ is a Wiener process along increasing $\lambda$.
With the exact score, these continuous-time SDEs share the prescribed marginals: the diffusion and its paired
score drift vary together~\citep{karras2022elucidating}.

\paragraph{Probability-flow ODE in log-SNR coordinates.}
Setting $\tilde{\beta}_{\lambda}=0$ and replacing $\epsilon_*$ with the learned prediction gives
\begin{equation}
d\mathbf{y}=-e^{-\lambda}\epsilon_{\theta}(\mathbf{y},\lambda,c)\,d\lambda.
\label{eq:app:dpm:pf_ode_lambda}
\end{equation}
Here $\epsilon_{\theta}(\mathbf{y},\lambda,c)$ is shorthand for the original denoiser evaluated at
$\mathbf{x}_t=\alpha_t\mathbf{y}$ and the corresponding timestep $t$.
Integrating the learned ODE from $s$ to $t$ gives
\begin{equation}
\mathbf{y}_t=\mathbf{y}_s-
\int_{\lambda_s}^{\lambda_t}e^{-\lambda}\epsilon_{\theta}(\mathbf{y}_{\lambda},\lambda,c)\,d\lambda.
\label{eq:app:dpm:exact_y}
\end{equation}
This is the noise-prediction representation of the diffusion ODE. DPM-Solver++ discretizes an equivalent
continuous-time data-prediction formulation~\citep{lu2022dpm_solver_plus_plus}.

\paragraph{Local propagation of auxiliary residual variances.}
At fixed current and stored states, the direct contribution of model-output perturbations to a solver step
is described by coefficients $w_{i,j}$, with $j\in\mathcal{J}_i$ indexing the predictions used in that step.
As in Sec.~\ref{sec:method:derivation}, we introduce independent auxiliary perturbations
$\boldsymbol{\eta}_j\sim\mathcal{N}(\mathbf{0},V_j\mathbf{I})$, where $V_j$ is the calibrated statistic
in Sec.~\ref{sec:method:stats}. Their propagated increment is
\begin{equation}
\delta\mathbf{y}_i^{\mathrm{aux}}=\sum_{j\in\mathcal{J}_i}w_{i,j}\boldsymbol{\eta}_j,
\label{eq:app:dpm:generic_update}
\end{equation}
with covariance
\begin{equation}
\operatorname{Cov}(\delta\mathbf{y}_i^{\mathrm{aux}})
=\left(\sum_{j\in\mathcal{J}_i}w_{i,j}^2V_j\right)\mathbf{I}.
\label{eq:app:dpm:var_injected}
\end{equation}
This local construction holds the state history fixed and omits dependence between successive solver
increments, which can share stored predictions.
For a diffusion strength $\tilde{\beta}_{\lambda,i}$ constant over the step, the diffusion term in
Eq.~\plaineqref{eq:app:dpm:sde_lambda} has increment covariance
$2\tilde{\beta}_{\lambda,i}\int_{\lambda_{i-1}}^{\lambda_i}e^{-2\lambda}\,d\lambda\,\mathbf{I}
=\tilde{\beta}_{\lambda,i}|\bar{\sigma}_{i-1}^2-\bar{\sigma}_i^2|\mathbf{I}$.
Matching this covariance to Eq.~\plaineqref{eq:app:dpm:var_injected} defines
\begin{equation}
\tilde{\beta}_{\lambda,i}
=\frac{\sum_{j\in\mathcal{J}_i}w_{i,j}^2V_j}
{|\bar{\sigma}_{i-1}^2-\bar{\sigma}_i^2|}.
\label{eq:app:dpm:beta_lambda}
\end{equation}
Here and below, $\alpha_i$, $\sigma_i$, and $\lambda_i$ abbreviate their values at $t_i$.

\paragraph{Correction scale and normalized update.}
The paired drift multiplier in Eq.~\plaineqref{eq:app:dpm:sde_lambda} motivates the choice
\begin{equation}
c_i\triangleq\tilde{\beta}_{\lambda,i}.
\label{eq:app:dpm:ci}
\end{equation}
We first compute the baseline quantized solver step $\mathbf{x}_i^{\mathrm{base}}$ from the current state
and stored predictions. We then rescale its full increment in normalized coordinates:
\begin{equation}
\begin{aligned}
\mathbf{y}_i^{\mathrm{base}}&=\mathbf{x}_i^{\mathrm{base}}/\alpha_i,\\
\mathbf{y}_i&=\mathbf{y}_{i-1}+(1+c_i)(\mathbf{y}_i^{\mathrm{base}}-\mathbf{y}_{i-1}),
\qquad \mathbf{x}_i=\alpha_i\mathbf{y}_i.
\end{aligned}
\label{eq:app:dpm:qdrift_update}
\end{equation}
This adapts the drift-rescaling principle to the discrete solver while retaining its baseline step computation.

\paragraph{Instantiation: DPM-Solver++(2M, midpoint).}
For noise prediction, DPM-Solver++ converts the quantized output to a data prediction:
\begin{equation}
\hat{\mathbf{x}}_{\theta,i}
=\frac{\mathbf{x}_i-\sigma_i\hat{\epsilon}_{\theta,i}}{\alpha_i}
=\mathbf{y}_i-\bar{\sigma}_i\hat{\epsilon}_{\theta,i}.
\label{eq:app:dpm:eps_to_xtheta}
\end{equation}
Let $h_i\triangleq\lambda_i-\lambda_{i-1}$ and $r_i\triangleq h_{i-1}/h_i$.
The second-order multistep midpoint update (Algorithm~2 of \citet{lu2022dpm_solver_plus_plus}) is
\begin{equation}
\begin{aligned}
\mathbf{x}_i^{\mathrm{base}}
&=\frac{\sigma_i}{\sigma_{i-1}}\mathbf{x}_{i-1}
-\alpha_i(e^{-h_i}-1)\hat{\mathbf{D}}_i,\\
\hat{\mathbf{D}}_i
&=\left(1+\frac{1}{2r_i}\right)\hat{\mathbf{x}}_{\theta,i-1}
-\frac{1}{2r_i}\hat{\mathbf{x}}_{\theta,i-2}.
\end{aligned}
\label{eq:app:dpm:2m_mid_eps}
\end{equation}
At fixed states, a perturbation $\Delta\epsilon_k$ changes the data prediction by
$-\bar{\sigma}_k\Delta\epsilon_k$. Its direct contribution to the normalized baseline step is therefore
\begin{equation}
\delta\mathbf{y}_i^{\mathrm{local}}=(e^{-h_i}-1)
\left[\left(1+\frac{1}{2r_i}\right)\bar{\sigma}_{i-1}\Delta\epsilon_{i-1}
-\frac{1}{2r_i}\bar{\sigma}_{i-2}\Delta\epsilon_{i-2}\right].
\label{eq:app:dpm:2m_mid_eps_dyq}
\end{equation}
The model-output perturbation coefficients used in Eq.~\plaineqref{eq:app:dpm:var_injected} are thus
\begin{equation}
w_{i,i-1}=(e^{-h_i}-1)\left(1+\frac{1}{2r_i}\right)\bar{\sigma}_{i-1},
\qquad w_{i,i-2}=-(e^{-h_i}-1)\frac{\bar{\sigma}_{i-2}}{2r_i}.
\label{eq:app:dpm:2m_mid_eps_weights}
\end{equation}
The resulting correction scale is
\begin{equation}
c_i=\tilde{\beta}_{\lambda,i}
=\frac{(e^{-h_i}-1)^2
\left[\left(1+\frac{1}{2r_i}\right)^2\bar{\sigma}_{i-1}^2V_{i-1}
+\left(\frac{1}{2r_i}\right)^2\bar{\sigma}_{i-2}^2V_{i-2}\right]}
{|\bar{\sigma}_{i-1}^2-\bar{\sigma}_i^2|}.
\label{eq:app:dpm:beta_lambda_2m_mid_eps}
\end{equation}
For first-order steps, including warmup, the same construction uses the single coefficient
$w_{i,i-1}=\bar{\sigma}_i-\bar{\sigma}_{i-1}$.
Finally, Eq.~\plaineqref{eq:app:dpm:qdrift_update} can be expressed in the original coordinates as
\begin{equation}
\mathbf{x}_i=(1+c_i)\mathbf{x}_i^{\mathrm{base}}
-c_i\frac{\alpha_i}{\alpha_{i-1}}\mathbf{x}_{i-1},
\label{eq:app:dpm:qdrift_2m_mid_eps}
\end{equation}
where $\mathbf{x}_i^{\mathrm{base}}$ is given by Eq.~\plaineqref{eq:app:dpm:2m_mid_eps} for a second-order step.

\section{Additional Experiments}

Unless otherwise noted, all supplementary experiments follow the same setup as in the main experiments.

\begin{table}[!t]
  \caption{Additional results on MJHQ-30K (milder quantization settings). Rows labeled W4A4 use W4A4 quantization for SVDQuant; ViDiT-Q rows use W8A8 quantization.}
  \label{tab:exp:main_additional}
  \centering

  \begin{tabular}{llccccc}
    \toprule
    Model & Method & FID$\downarrow$ & CLIP$\uparrow$ & PSNR$\uparrow$ & LPIPS$\downarrow$ & SSIM$\uparrow$ \\
      \midrule
    FLUX.1-dev & FP & 20.68 & 25.80 & -- & -- & -- \\
    (SVDQuant W4A4) & Quantized & \textbf{20.49} & 25.68 & 22.95 & 0.194 & 0.830 \\
    & Q-Drift & 20.55 & 25.68 & 22.98 & 0.193 & 0.830 \\
    \midrule
    FLUX.1-schnell & FP & 19.18 & 26.55 & -- & -- & -- \\
    (SVDQuant W4A4) & Quantized & \textbf{18.62} & 26.41 & 18.44 & 0.251 & 0.728 \\
    & Q-Drift & 18.69 & 26.44 & 18.42 & 0.251 & 0.727 \\
    \midrule
    SDXL & FP & 17.20 & 27.40 & -- & -- & -- \\
    (SVDQuant W4A4) & Quantized & \textbf{17.05} & 27.27 & 20.41 & 0.270 & 0.728 \\
    & Q-Drift & 17.17 & 27.29 & 20.41 & 0.270 & 0.727 \\
    \midrule
    SDXL-Turbo & FP & 24.77 & 26.39 & -- & -- & -- \\
    (SVDQuant W4A4) & Quantized & 24.75 & 26.37 & 18.44 & 0.242 & 0.656 \\
    & Q-Drift & \textbf{24.51} & 26.42 & 12.85 & 0.549 & 0.380 \\
    \midrule
    PixArt-$\Sigma$ & FP & 16.52 & 26.71 & -- & -- & -- \\
    (SVDQuant W4A4) & Quantized & \textbf{19.61} & 26.37 & 16.75 & 0.360 & 0.629 \\
    & Q-Drift & 19.65 & 26.36 & 16.71 & 0.368 & 0.619 \\
    \midrule
    PixArt-$\Sigma$ & FP & 16.52 & 26.71 & -- & -- & -- \\
    (ViDiT-Q W8A8) & Quantized & 15.67 & 26.65 & 21.11 & 0.176 & 0.774 \\
    & Q-Drift & \textbf{15.62} & 26.67 & 21.08 & 0.176 & 0.775 \\
    \midrule
    Sana & FP & 15.98 & 27.30 & -- & -- & -- \\
    (SVDQuant W4A4) & Quantized & 15.74 & 27.23 & 17.27 & 0.260 & 0.637 \\
    & Q-Drift & \textbf{15.68} & 27.24 & 17.16 & 0.270 & 0.626 \\
      \bottomrule
  \end{tabular}

\end{table}

\subsection{Additional Results (Milder Quantization Settings)}
\label{sec:app:main_additional}

Table~\ref{tab:exp:main_additional} reports additional results under milder quantization settings for SVDQuant and
includes an additional PTQ method, ViDiT-Q~\citep{zhao2025viditq}.
Under these milder settings, the quantized baseline can already match or outperform full precision in FID, leaving less
headroom for improvement.
Accordingly, the gains from Q-Drift are mixed across settings.

\subsection{Empirical Examination of the Modeling Assumptions}
\label{sec:app:assumption_validation}

We examine selected distributional and covariance properties underlying the modeling assumptions in
Sec.~\ref{sec:prelim:joint-gauss}--Sec.~\ref{sec:prelim:joint-simplify} on SDXL with SVDQuant W3A4, using
a larger set of 5K paired full-precision and quantized runs collected for this analysis; the correction factors in the
main experiments are fitted on the 1K-prompt calibration set (Sec.~\ref{sec:exp:setup}).
We examine three representative timesteps, chosen as the nearest discrete indices to
$0.9T$, $0.5T$, and $0.1T$.

\begin{figure}[!t]
  \centering
  \includegraphics[width=\textwidth]{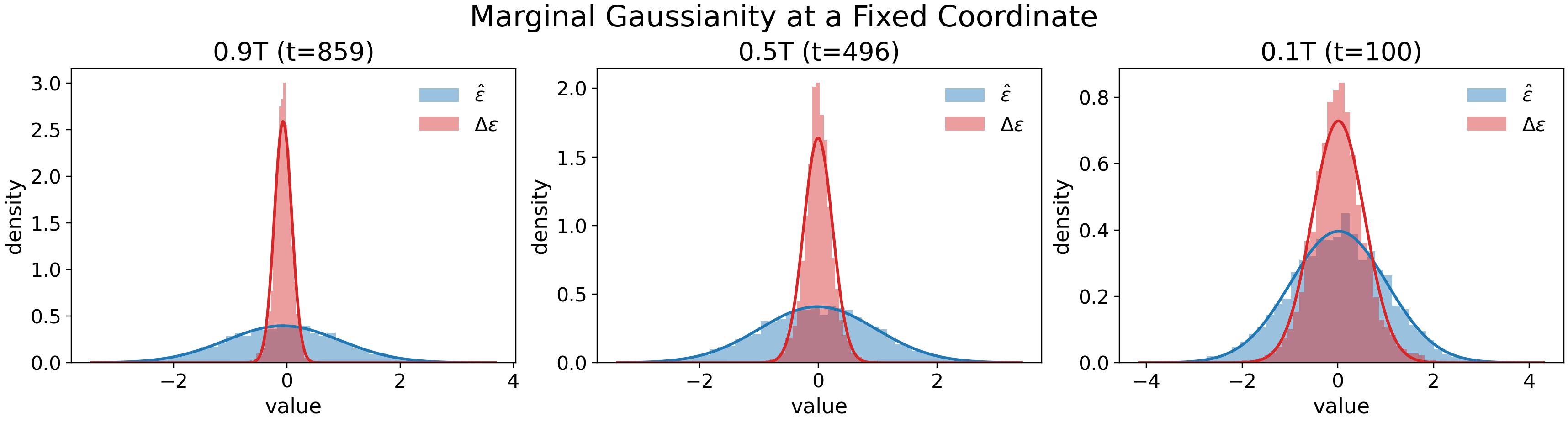}

  \includegraphics[width=\textwidth]{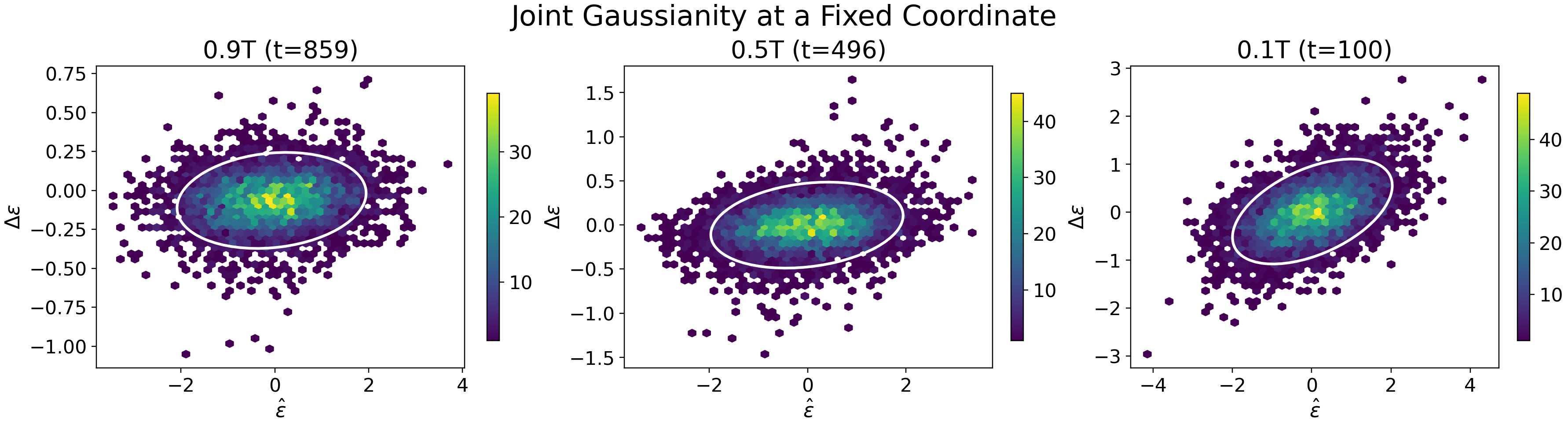}
  \caption{Empirical validation of marginal and joint Gaussianity on SDXL (SVDQuant W3A4).
  Top: Histograms of $\hat{\epsilon}_{\theta}^{(t)}$ and $\Delta\epsilon_{\theta}^{(t)}$ at a latent coordinate,
  overlaid with fitted Gaussian curves.
  Bottom: The corresponding 2D joint densities with fitted Gaussian ellipses.
  All 1D and 2D visualizations are shown at the same latent coordinate, $(c,h,w)=(0,64,64)$.}
  \label{fig:app:assumption_marginal_joint}
\end{figure}

\paragraph{Marginal and joint Gaussianity.}
This experiment directly probes the joint Gaussian modeling assumption introduced in
Sec.~\ref{sec:prelim:joint-gauss}.
Fig.~\ref{fig:app:assumption_marginal_joint} shows that the marginal distributions of
$\hat{\epsilon}_{\theta}^{(t)}$ and $\Delta\epsilon_{\theta}^{(t)}$ at the selected coordinate are
well-approximated by fitted Gaussians across timesteps.
The corresponding joint densities are approximately elliptical, with the fitted Gaussian ellipse capturing the dominant
orientation and spread of the samples.
These observations support the timestep-wise Gaussian approximation at the examined coordinate.

\begin{figure}[!t]
  \centering
  \includegraphics[width=\textwidth]{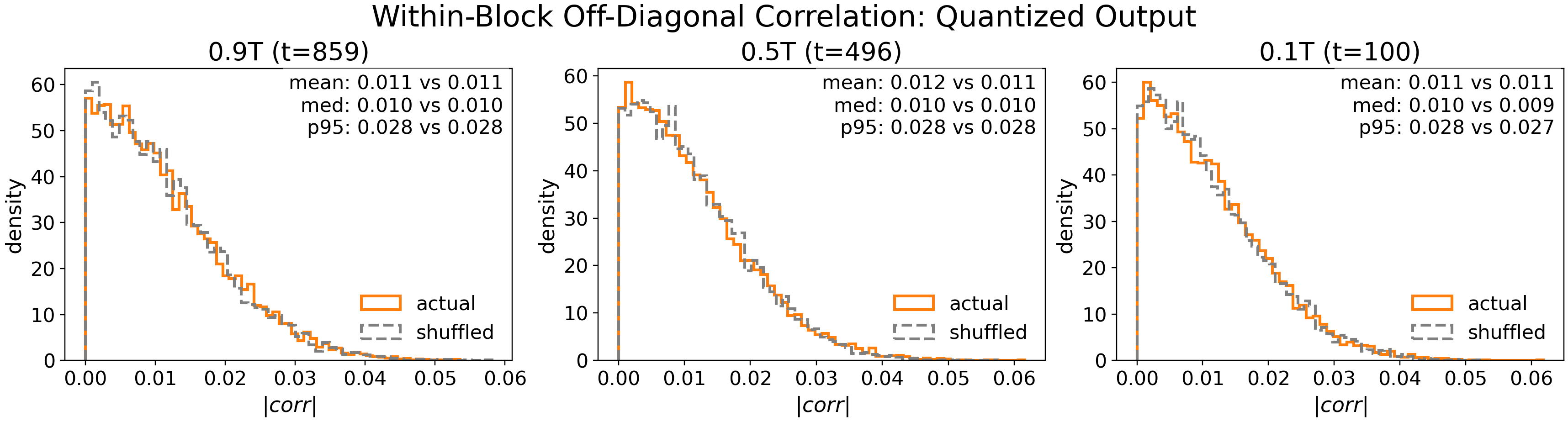}

  \includegraphics[width=\textwidth]{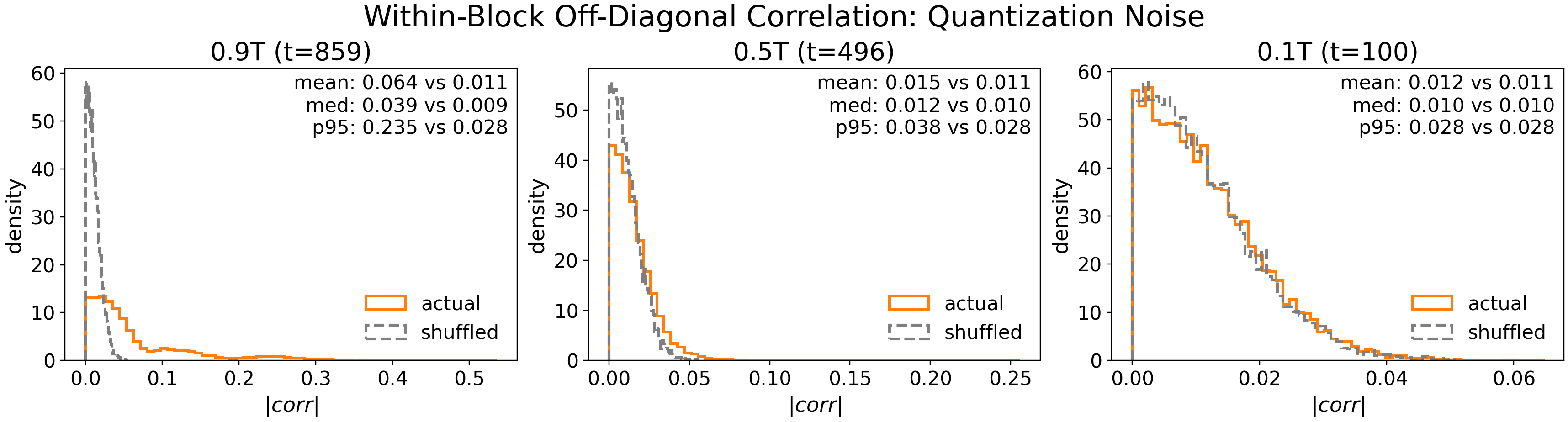}

  \includegraphics[width=\textwidth]{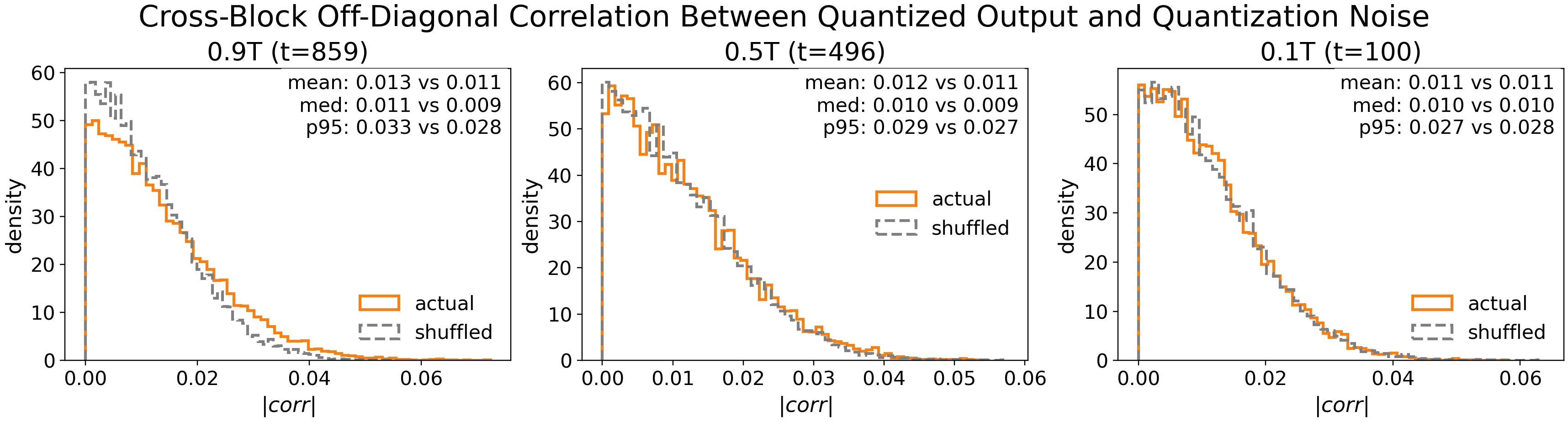}
  \caption{Empirical validation of the diagonal-covariance simplification.
  For each selected timestep, we compare the distribution of absolute correlations from 10,000 random off-diagonal pairs,
  with each correlation estimated over 5,000 calibration samples, against a shuffled baseline.
  The shuffled baseline is obtained by randomly permuting a variable in each pair.
  In each panel, the inset reports the mean, median (med), and 95th percentile (p95) of the absolute-correlation
  distribution, listed as actual vs shuffled.
  Top: off-diagonal entries of $\Sigma_{\hat{\epsilon}\hat{\epsilon}}^{(t)}$.
  Middle: off-diagonal entries of $\Sigma_{\Delta\Delta}^{(t)}$.
  Bottom: off-diagonal cross-block entries of $\Sigma_{\hat{\epsilon}\Delta}^{(t)}$ for $i\neq j$.}
  \label{fig:app:assumption_corr}
\end{figure}

\paragraph{Diagonal covariance as a tractable approximation.}
The more restrictive assumption in Sec.~\ref{sec:prelim:joint-simplify} is the diagonal approximation to the
covariance blocks $\Sigma_{\hat{\epsilon}\hat{\epsilon}}^{(t)}$, $\Sigma_{\Delta\Delta}^{(t)}$, and
$\Sigma_{\hat{\epsilon}\Delta}^{(t)}$.
This simplification is practically necessary since with only 5K calibration samples, directly estimating a covariance
block over a latent tensor of size $C\times H\times W = 4\times 128\times 128 = 65{,}536$ would be statistically unstable.
In this regime, the ambient dimensionality already exceeds the number of available samples, so the resulting sample
covariance would be severely underdetermined and its off-diagonal structure would be poorly estimated.

To assess whether the off-diagonal entries are negligible in practice, we compare the empirical distribution of
absolute correlations from random off-diagonal pairs against a shuffled baseline.
The shuffled baseline is obtained by randomly permuting a variable in each pair,
which preserves the marginal distribution of each element while removing pairwise dependence.
This comparison is necessary because finite-sample estimation produces nonzero empirical correlations even under
independence. Accordingly, the shuffled baseline estimates the finite-sample noise floor associated with the 5K
calibration set. Agreement with this baseline indicates little resolved pairwise dependence among the sampled
entries at the available sample size.
Fig.~\ref{fig:app:assumption_corr} reports three complementary checks:
off-diagonal entries of $\Sigma_{\hat{\epsilon}\hat{\epsilon}}^{(t)}$,
off-diagonal entries of $\Sigma_{\Delta\Delta}^{(t)}$,
and off-diagonal cross-block entries of $\Sigma_{\hat{\epsilon}\Delta}^{(t)}$ for $i\neq j$.

Two observations are important.
First, the off-diagonal entries of $\Sigma_{\hat{\epsilon}\hat{\epsilon}}^{(t)}$ and the off-diagonal cross-block
entries of $\Sigma_{\hat{\epsilon}\Delta}^{(t)}$ are generally close to the shuffled baseline.
Second, the off-diagonal entries of $\Sigma_{\Delta\Delta}^{(t)}$ exhibit non-negligible dependence at the earliest,
highest-noise timestep, while the gap to the shuffled baseline becomes much smaller at later timesteps.
Therefore, the diagonal covariance model should be viewed as a calibration-friendly approximation rather than an exact description.

\begin{table}[!t]
  \centering
  \caption{\textbf{Distributional metrics and statistical uncertainty} on MJHQ-30K, computed on the same
  generated image sets used in Table~\ref{tab:exp:main}. KID is in $\times 10^{3}$ scale and brackets denote
  $95\%$ paired bootstrap intervals. A $\Delta$ interval lying entirely below zero indicates an improvement supported by
  the sample.}
  \label{tab:app:kid_ci}

  \begin{tabular}{llccc}
    \toprule
    Model & Method & FID$\downarrow$ & $\Delta$FID 95\% CI & KID$\downarrow$ \\
    \midrule
    FLUX.1-dev & FP & 20.68 & -- & 3.44 \\
    (SVDQuant W3A4) & Quantized & 24.14 & -- & 5.67 \\
     & Q-Drift & 24.06 & [$-$0.30, 0.09] & 5.59 \\
    \midrule
    FLUX.1-schnell & FP & 19.18 & -- & 2.66 \\
    (SVDQuant W3A4) & Quantized & 23.10 & -- & 5.03 \\
     & Q-Drift & 22.13 & [$-$1.18, $-$0.84] & 4.52 \\
    \midrule
    SDXL & FP & 17.20 & -- & 1.49 \\
    (SVDQuant W3A4) & Quantized & 31.73 & -- & 9.36 \\
     & Q-Drift & 30.69 & [$-$1.38, $-$0.70] & 8.65 \\
    \midrule
    SDXL-Turbo & FP & 24.77 & -- & 6.76 \\
    (SVDQuant W3A4) & Quantized & 29.17 & -- & 8.75 \\
     & Q-Drift & 27.64 & [$-$1.83, $-$1.24] & 7.80 \\
    \midrule
    SDXL-Turbo & FP & 24.77 & -- & 6.76 \\
    (MixDQ W4A8) & Quantized & 28.37 & -- & 8.32 \\
     & Q-Drift & 27.38 & [$-$1.49, $-$0.50] & 7.86 \\
    \midrule
    PixArt-$\Sigma$ & FP & 16.52 & -- & 1.86 \\
    (SVDQuant W3A4) & Quantized & 48.85 & -- & 17.47 \\
     & Q-Drift & 44.06 & [$-$5.24, $-$4.26] & 14.47 \\
    \midrule
    Sana & FP & 15.98 & -- & 1.51 \\
    (SVDQuant W3A4) & Quantized & 17.32 & -- & 2.07 \\
     & Q-Drift & 16.54 & [$-$0.96, $-$0.57] & 1.71 \\
    \bottomrule
  \end{tabular}

\end{table}

\section{Distributional Metrics and Statistical Uncertainty}
\label{sec:app:kid_ci}
FID is a point estimate computed on a finite sample, so a raw FID gap does not by itself show that one
sampler produces a closer distribution than another. Table~\ref{tab:app:kid_ci} therefore reports two further
quantities alongside FID.

The first is KID, an unbiased estimator of squared polynomial-kernel MMD~\citep{binkowski2018demystifying},
which provides an additional measure of distributional fidelity. Its expectation is independent of sample size,
although its sampling variability depends on it. We compute KID from $100$ random subsets of $1{,}000$ images
and report it in $\times 10^{3}$ scale.

The second is a paired $95\%$ percentile bootstrap interval for
$\Delta\mathrm{FID}=\mathrm{FID}(\text{Q-Drift})-\mathrm{FID}(\text{Quantized})$.
We use $1{,}000$ resamples, sharing the sampled indices of aligned generated images between the two methods
and using a common resampled reference set within each replicate.
The interval endpoints are the $2.5$th and $97.5$th percentiles of the bootstrap differences.
Pairing retains the dependence between the two FID estimates, but does not guarantee cancellation of their
finite-sample biases.

Q-Drift lowers KID in every setting. Six of the seven $\Delta\mathrm{FID}$ intervals lie entirely below zero;
the exception is FLUX.1-dev, whose FID gain is the smallest of all settings and is not resolved at this sample
size. No interval in the table lies above zero.

\section{Prompts for the Qualitative Visual Comparisons}
\label{sec:app:visual_prompts}

The prompts corresponding to the five rows in Fig.~\ref{fig:exp:visual_sdxl_w3a4}
are listed below in row order.

\noindent Row 1: ``ping\"uino con auriculares puestos subiendo las aletas en la
piscina del hotel monge tibetano poniendole unos auriculares a un tigre blanco
ultra realistic 3d photo v5.''

\noindent Row 2: ``Futuristic Lace dress, Fashion Concept, led lights. Sexy
attractive pose. Hotpants. Beautiful, cinematic lighting, 8K, sharp details.
Crystal clear feel, High quality, Fineart print, HQ. Tron vibes. Futuristic
background. colourful. Hypermaximalist, hyperdetailed, awardwinning fashion
photography, professional colour grading, clean sharp focus. Clean details.''

\noindent Row 3: ``a moment in time of a stunning, cute couple, grinding on each
other.''

\noindent Row 4: ``wellbuilt, beefy, stout, heavyset, mature man, wide shot,
Realistic editorial photography session featuring a stocky very handsome in his
fourties with light brown hair and fair skin, handsome, attractive, interesting
background, art by Jean Baptiste Monge, Lisa frank, Boris Vallejo, shot
photography by Wes Anderson, Nikon Photography Symmetrical Frontal Portrait,
High Definition, 8K, 200mm.''

\noindent Row 5: ``an image of a shop owner for a sustainable business.''

\clearpage

\end{document}